\documentclass[10pt,journal,compsoc]{IEEEtran}
\ifCLASSOPTIONcompsoc
  \usepackage[nocompress]{cite}
\else
  \usepackage{cite}
\fi
\usepackage{graphicx}

\usepackage{amsmath}
\usepackage{amsfonts}
\usepackage{amssymb}
\usepackage{algorithm}
\usepackage{algorithmic}
\ifCLASSOPTIONcompsoc
  \usepackage[caption=false,font=footnotesize,labelfont=sf,textfont=sf]{subfig}
\else
  \usepackage[caption=false,font=footnotesize]{subfig}
\fi

\usepackage{booktabs}
\usepackage{hyperref}

\begin{document}
\title{CenGCN: Centralized Convolutional Networks with Vertex Imbalance for Scale-Free Graphs}

\author{Feng~Xia,~\IEEEmembership{Senior Member,~IEEE,}
        Lei~Wang,	
        Tao~Tang,
        Xin~Chen,
        Xiangjie~Kong,~\IEEEmembership{Senior Member,~IEEE,}
        Giles~Oatley,
       and Irwin~King,~\IEEEmembership{Fellow,~IEEE}% <-this % stops a space
\thanks{This work was partially supported by the National Natural Science Foundation of China (62072409) and Zhejiang Provincial Natural Science Foundation (LR21F020003).}
\IEEEcompsocitemizethanks{\IEEEcompsocthanksitem F. Xia, T. Tang, and G. Oatley are with School of Engineering, IT and Physical Sciences, Federation University Australia, Ballarat, VIC 3353, Australia. 
% E-mail: see http://www.michaelshell.org/contact.html
\IEEEcompsocthanksitem L. Wang and X. Chen are with School of Software, Dalian University of Technology, Dalian 116620, China.
\IEEEcompsocthanksitem X. Kong is with College of Computer Science and Technology, Zhejiang University of Technology, Hangzhou 310023, China.
\IEEEcompsocthanksitem I. King is with Department of Computer Science and Engineering, The Chinese University of Hong Kong, Hong Kong.}
%% <-this % stops an unwanted space
\thanks{Corresponding author: Xiangjie Kong; email: xjkong@ieee.org.}}


%\markboth{Journal  \LaTeX\ Class Files,~Vol.~00, No.~0, January~2020}%
% \markboth{IEEE Transactions on Knowledge and Data Engineering,~Vol.~00, No.~0, June~2020}%
\markboth{IEEE Transactions on Knowledge and Data Engineering}%
{First Author \MakeLowercase{\textit{et al.}}: Centralized Convolutional Networks with Vertex Imbalance for Scale-Free Graphs}



\IEEEtitleabstractindextext{%
\begin{abstract}
Graph Convolutional Networks (GCNs) have achieved impressive performance in a wide variety of areas, attracting considerable attention. The core step of GCNs is the information-passing framework that considers all information from neighbors to the central vertex to be equally important. Such equal importance, however, is inadequate for scale-free networks, where hub vertices propagate more dominant information due to vertex imbalance. In this paper, we propose a novel centrality-based framework named CenGCN to address the inequality of information. This framework first quantifies the similarity between hub vertices and their neighbors by label propagation with hub vertices. Based on this similarity and centrality indices, the framework transforms the graph by increasing or decreasing the weights of edges connecting hub vertices and adding self-connections to vertices. In each non-output layer of the GCN, this framework uses a hub attention mechanism to assign new weights to connected non-hub vertices based on their common information with hub vertices. We present two variants CenGCN\_D and CenGCN\_E, based on degree centrality and eigenvector centrality, respectively. We also conduct comprehensive experiments, including vertex classification, link prediction, vertex clustering, and network visualization. The results demonstrate that the two variants significantly outperform state-of-the-art baselines.
\end{abstract}

}

\maketitle

\IEEEdisplaynontitleabstractindextext

% \fi
%\IEEEpeerreviewmaketitle

\IEEEraisesectionheading{\section{Introduction}\label{sec:introduction}}

%\IEEEPARstart{T}{his} demo file is intended to serve as a ``starter file''
%for IEEE Computer Society journal papers produced under \LaTeX\ using
%IEEEtran.cls version 1.8b and later.
%% You must have at least 2 lines in the paragraph with the drop letter
%% (should never be an issue)
%I wish you the best of success.

\IEEEPARstart{T}{he} graph, as an abstract data type, can represent the complex relationships between objects in many real-world networks. Representative networks include social networks \cite{stadtfeld2019integration}, biological networks \cite{kovacs2019network}, and academic networks \cite{xia2017big}. Numerous studies \cite{cui2018survey,cai2018comprehensive,liu2018artificial,xia2021chief} demonstrate the possibilities of extracting rich information from graph-structured data, thereby realizing many practical applications, including vertex classification and link prediction. However, how to extract useful information from these data remains a challenging issue and is thus worthy of exploration in depth.

Recently, extensive studies \cite{kipf2017semi,li2018deeper,defferrard2016convolutional,velickovic2019deep,Xia2021TAI} have shown that Graph Convolutional Networks (GCNs) are powerful tools for handling graph-structured data and for a wide spectrum of graph-based applications, from recommender systems \cite{ying2018graph,wang2020Multi} to knowledge graphs \cite{nathani2019learning,kong2020gene}. Existing GCNs adopt an information-passing framework \cite{gilmer2017neural}, where each vertex aggregates information from its immediate neighbors and itself, and considers information from different vertices equally important. However, such equal importance is counterintuitive when different neighbors pass information with different influence to the central vertex. For instance, a person could connect to both friends and work colleagues in social networks where vertices denote persons. When recognizing one's workplace, the information from colleagues is more relevant than that from friends. In this instance, we need to weight more highly the influence of colleagues. Equally aggregating information, however, fails to capture this differentiation between vertices.

Many complex networks in the real world, such as the Internet and social networks, are scale-free networks \cite{faloutsos1999power}. We find in these networks an inequality of information from different vertices because of vertex imbalance. The scale-free property, one of the fundamental macroscopic structures of networks, dictates that the vertex degrees follow a power-law distribution: the probability distribution decreases as the vertex degree increases, with a long tail tending to zero. Therefore, significant vertex imbalance appears in a scale-free network, and only a few vertices are of high degree and regarded as hub vertices or simply hubs. The majority of vertices linking to a high-degree vertex are, however, of low degree, and not highly connected. In scale-free networks, a vertex with a high degree is usually the hub of a community, and the information it contains is more influential than that from vertices with low degrees. For instance, in social media, a  celebrity that has a great number of followers can spread more news than a less prestigious person. In this case, if we wish to tap into a users' interests, it would be helpful to consider links to celebrities. Additionally, when modeling the diffusion, more weight should be given to edges connecting hub vertices \cite{smith2013spectral}.

For scale-free networks, it is possible to implement new GCNs that capture this differentiation in information passing between vertices. Since vertex centrality can be considered a useful measure of the importance of individual vertices \cite{gu2018rare}, we propose a centrality-based Graph Convolution Network. In this paper, we consider various centrality measurements, rather than solely focusing on vertex degree. The degree of a vertex is a centrality measurement. The proposed GCNs allow hub vertices, selected by centrality indices, to pass more information. Thus, each vertex receives more information from the hub vertices.

When designing centrality-based GCNs, we need to address the issue that a vertex with a higher centrality index may be linked to similar vertices and possibly to dissimilar vertices due to its high prestige and popularity \cite{feng2018representation}. For example, in academia, some distinguished scholars can collaborate with others from outside their research laboratories. In social networks, a celebrity receives a great number of followers, but most of them may have totally different backgrounds. 

The relationship between two connected but dissimilar vertices needs to be weakened. When aggregating information, neighboring information with similar features is aggregated to reinforce the correct features to facilitate downstream tasks such as classification, while the opposite effect is achieved if dissimilar information is aggregated. Therefore, different weights are required. To address the issue detailed above, we consider the similarity between vertices in the underlying network structure. We use random walk computation in order to calculate this similarity \cite{perozzi2014deepwalk,grover2016node2vec,yan2019constrained,Xia2019TETCIrandom}. Subsequently, a label propagation algorithm is applied over hub vertices to quantify the similarity between vertices and their hub neighbors. Through these quantified similarities and vertex centrality indices, we propose a graph transformation method to increase or decrease the weights of edges connecting hub vertices and to add self-connections to vertices. In the transformed graph, the influence from hub vertices to their similar neighbors is strengthened, and the influence to dissimilar neighbors is weakened.

In the transformed graph, vertices are influenced by their hub neighbors. It is possible that some neighbors of a vertex are non-hub and these neighbors will also have an effect. Therefore, we propose a hub attention mechanism that passes information between non-hub vertices that share common hubs. This attention mechanism is faster and has fewer parameters to be learned than the previous graph attention mechanism \cite{velickovic2018graph}.

\begin{figure}[htbp]
	\centering
	\includegraphics[width=2.5in]{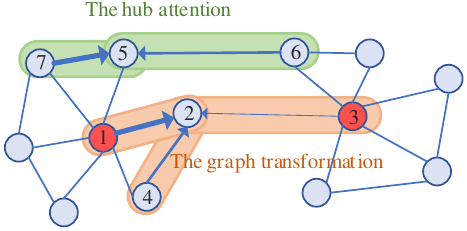}
	\caption{An illustrative example, where $v_1$ and $v_3$ are hub vertices, and vertex $v_2$ is highly similar to hub vertex $v_1$, but not to hub vertex $v_3$. The graph transformation is proposed to handle hub neighbors, and the hub attention is proposed to handle non-hub neighbors.
	}
	\label{figure:illustration}
\end{figure}

Fig. \ref{figure:illustration} presents an example to illustrate how the graph transformation and the hub attention work. Two hub vertices, $v_1$ and $v_3$, are marked by red. In the network structure, $v_2$ are similar with $v_1$ but dissimilar with $v_3$. When the information of $v_1$, $v_3$ and $v_4$ runs into $v_2$, the graph transformation gives a higher weight to the edge $e_{12}$, a lower weight to the edge $e_{23}$, and the original weight to the edge $e_{24}$. $v_5$ shares the same hub with $v_7$ but no same hub with $v_6$. When the information of $v_7$ and $v_6$ runs into $v_5$, the hub attention mechanism gives a higher weight to the edge $v_{57}$ and a lower weight to the edge $v_{56}$. Note that self-connections are omitted for clarification of how to handle neighbors.

We name our overall framework as CenGCN. In this paper, we present two variants of CenGCN, CenGCN\_D and CenGCN\_E, based on degree centrality and eigenvector centrality, respectively. To evaluate their performances, we conducted four experiments, vertex classification, link prediction, vertex clustering, and network visualization, on five datasets.  These experiments show that the two variants outperform state-of-the-art baselines, even by 70.1\% on vertex classification. Though the scale-free property is based on vertex degree, the finding that CenGCN\_E, based on eigenvector centrality, achieves excellent performance inspires us to utilise more centrality measurements for GCNs and additional graph-based methods. Further, experiments also show that CenGCN\_D and CenGCN\_E exhibit a greater performance over GCNs as the network becomes deeper. Thus, an observation  derived from this study is that we should explore vertex imbalance and unequal information by vertex centrality on deeper GCNs.

The contributions of this paper can be summarized as follows:
\begin{itemize}
	\item We propose a framework named CenGCN for scale-free networks. This framework effectively addresses the unequal importance of information from different vertices.
	\item We propose using label propagation to quantify the similarity between hub vertices and their neighbors, a graph transformation method that captures the influence of hub vertices by vertex centrality, and a hub attention mechanism that assigns new weights to non-hub neighbors by the same hubs.
	\item We present two variants of CenGCN, namely CenGCN\_D and CenGCN\_E. Extensive experiments show that these two variants outperform state-of-the-art baselines as well as deeper GCNs.
\end{itemize}

The remainder of this paper is structured as follows. In Section 2, we give an overview of related work. In Section 3, we introduce preliminaries related to this study and then introduce the CenGCN framework in Section 4. We present extensive experiments to verify the efficacy of this framework in Section 5. Finally, we conclude this paper in Section 6.

\section{Related Work}
In this section, we review related work about GCNs and scale-free networks.
\subsection{Graph Convolutional Networks}
Shuman et al. introduce a convolution operation on graph-structured data using the Fourier basis for signal processing in the paper \cite{shuman2013emerging}. Based on this study \cite{shuman2013emerging}, Bruna et al. \cite{bruna2014spectral} define a convolution for graphs from the spectral domain using the graph Laplacian. Theoretical analysis shows that this definition of the convolution operation on graphs can mimic certain geometric properties of Convolutional Neural Networks (CNNs) \cite{bronstein2017geometric}. A significant limitation of this convolution is that the decomposition of the Laplacian matrix is not scalable to large-scale graphs. To solve the efficiency problem, Defferrard et al. \cite{defferrard2016convolutional} propose a $K$-localized spectral filter represented by $K$-order polynomials in the Laplacian. Through the Chebyshev expansion, we can recursively and fast compute the localized filter. With $K$ set to 1, Kipf et al. \cite{kipf2017semi} consider only the first-order neighbors and define a layer-wise propagation rule. The form of this propagation rule is a first-order approximation of localized spectral filters defined in the paper \cite{defferrard2016convolutional}. The first-order GCNs yield a fascinating performance, and many methods have been proposed to improve it.

Graph Attention Networks (GATs) \cite{velickovic2018graph} observe that the contributions from neighbors to the central vertex are unequal and adopt attention mechanisms to learn the relative weights between two connected vertices. Furthermore, GATs employ multi-head attention to stabilize the learning process of self-attention. Hierarchical Graph Convolutional Networks (H-GCN) \cite{hu2019hierarchical} address the failure of GCNs to obtain adequate global information. They repeatedly aggregate structurally similar nodes to hypernodes and then refine the coarsened graph to the original to restore the representation for each node in order to increase the receptive field of each vertex. GCNs are designed for semi-supervised learning, and to extend to unsupervised learning DGI \cite{velickovic2019deep} presents a general approach for learning vertex  representations within graph-structured data by maximizing mutual information between patch representations and corresponding high-level summaries of graphs.
\subsection{Scale-Free Networks}
The scale-free property describes how vertex degrees follow a power-law distribution in some networks, such as the Internet \cite{faloutsos1999power}. The study presented in \cite{newman2005power} reviews some of the empirical evidence for the existence of power-law forms and the theories proposed to explain them. Clauset et al. \cite{clauset2009power}  present a principled statistical framework for discerning and quantifying power-law behavior in empirical data. To define precisely the scale-free graphs, the study \cite{li2005towards} provides one possible measure of the extent to which a graph is scale-free. Considering the scale-free property of real-world networks, Jo et al. \cite{jo2019realgraph} propose a single-machine based graph engine equipped with the hierarchical indicator and the block-based workload allocation.

Despite many studies about the scale-free property, existing GCNs-based methods have not yet considered it.
Recently,  Feng et al. \cite{feng2018representation} proposed a principle for scale-free property preserving network embedding algorithms. Feng et al's study has three significant differences to our study: (i) we believe that those neighbors of a high-degree vertex contain both similar vertices and dissimilar vertices with it; Feng et al's study only assumes a high-degree vertex is dissimilar to its neighbors; (ii) we either reward or punish hub vertices with high centrality indices; Feng et al's study punishes vertices with high degrees; (iii) we learn vector representations using GCNs, or non-linear deep models; Feng et al's study uses spectral cluster and random walk, both of which are linear.

\section{Preliminaries}
In this section, we introduce the preliminaries related to this study, including the definitions of graphs and GCNs.
\subsection{Graph}
We consider a graph $G=(V,E)$, where $V=\{v_1,v_2,...,v_n\}$ is the vertex set containing $n$ vertices, and $E=\{e_{ij}\}_{1\leq i,j \leq n}$ is the edge set. If an undirected edge $e_{ij}$ exists between  $v_i$  and $v_j$,  $e_{ij} \in E$. We define  the adjacency matrix of $G$ as $A \in
\mathbb{R}^{n\times n}$, where $A_{ij}=1$ if $e_{ij} \in E$, and  $A_{ij}=0$ otherwise. We use $D$ to denote the degree diagonal matrix with $D_{ii}=\sum_{j}A_{ij}$. For the considered graph, we define a feature matrix $X \in \mathbb{R}^{n\times m}$, where the $i_{th}$ row $X_i$ is $v_i$'s features, and $m$ is the number of the features. Each vertex $v_i$ has a neighbor set $N_i$. If $A_{ij}=1$, $v_j\in N_i$.

A graph has scale-free property if its vertex degrees follow a power-law distribution. In this type of graph, only a few vertices are of high degree and called hub vertices or hubs. The majority of vertices connected to a high-degree vertex are of low degree and are not likely to be connected to each other. Formally, the probability density function of the vertex degree $D_{ii}$ has the following form:
\begin{equation}
P_{D_{ii}}(d)=Cd^{-\alpha}, \alpha>1,d>d_{min}>0,
\end{equation}
where $\alpha$ is the exponent parameter, and $C$ is the normalization term. The power-law form only applies to vertices with degrees greater than a certain minimal value $d_{min}$ \cite{clauset2009power}.

In graph theory, centrality has been extensively studied. A vertex with a higher centrality index usually is more influential and has greater prestige. To measure vertices' centrality indices, a number of methods have been put forward \cite{koschutzki2005centrality}. A well-known measure is degree centrality \cite{freeman1978centrality}, which regards $D_{ii}$ as the index of $v_i$'s centrality. Another popular measure is eigenvector centrality \cite{bonacich1972factoring}. It is defined as the principal eigenvector of the adjacency matrix defining the network. Other centrality measurements include closeness centrality,  betweenness, information centrality, flow betweenness and others \cite{borgatti2005centrality}.
\subsection{Graph Convolutional Networks (GCNs)}
The convolution operation on graph $G$ is defined in the Fourier domain:
\begin{equation}
y=g_\theta(L)x=g_\theta(U\Lambda U^T)x=Ug_\theta(\Lambda) U^Tx,
\end{equation}
where $L=I_n-D^{-\tfrac{1}{2}}AD^{-\tfrac{1}{2}}$ is the normalized Laplacian. $I_n$ is the identity matrix. $\Lambda$ and $U$ are the diagnonal matrix of eigenvalues and the matrix of eigenvectors of $L$, respectively. $x$ is the input signal, and $y$ is the filtered signal. $g_\theta (\Lambda)$ is the parameterized filter defined by \cite{defferrard2016convolutional} as a $K^{th} $ order polynomial:
\begin{equation}
g_\theta (\Lambda)=\sum_{k=0}^{K-1}\theta_k\Lambda^k,
\end{equation}
where the parameter $\theta_k$ is the polynomial coefficient. To circumvent the multiplication with Fourier basis $U$ that has $O(n^2)$ operations, Defferrard et al. \cite{defferrard2016convolutional}  adopt the Chebyshev polynomial $T_k(x)$ of order $k$ computed by the recurrence relation $T_k(x)=2xT_{k-1}(x)-T_{k-2}(x)$ with $T_0=1$ and $T_1=x$. Thus, the filter is parameterized as:
\begin{equation}
g_\theta(\Lambda)=\sum_{k=1}^{K-1}\theta_kT_k(\hat{\Lambda}),
\end{equation}
where $\hat{\Lambda}=2\Lambda/\lambda_{max}-I_n$ is a diagonal matrix of scaled eigenvalues, and  $\lambda_{max}$ is the maxmium eigenvalue of $L$.  The filtering operation can then be written as $y=g_\theta(L)x=\sum_{k=0}^{K-1}\theta_kT_k(\hat{L})x$, where $\hat{L}=2L/\lambda_{max}-I_n$ is the scaled Laplacian.

Further, let $\lambda_{max}=2$ and $K=1$, we can reach the GCNs \cite{kipf2017semi} defined by a layer-wise convolutional operation with the following layer-wise propagation rule:
\begin{equation}
H^{k+1}=\sigma(\tilde{D}^{-\tfrac{1}{2}} \tilde{A} \tilde{D}^{-\tfrac{1}{2}}H^kW^k) \label{conv1},
\end{equation}
Here, $\tilde{A}=A+I$ is the adjacency matrix of the undirected graph $G$ with added self-connections. $I$ is the identity matrix. $\tilde{D}$ is the degree diagonal matrix affiliated to $\tilde{A}$.  $W^k$ is the learnable weight in $k_{th}$ layer. $\sigma(\cdot)$ is an activation function, such as $ReLU(\cdot)$. $H^k$ is the input in the $k_{th}$ layer. We set $H^0=X$.
From Eq. \eqref{conv1}, we can see that GCNs can  be understood as special cases of
a simple differentiable information-passing framework \cite{gilmer2017neural}, i.e., aggregating information from neighbors and itself. One alternative propagation rule often used \cite{schlichtkrull2018modeling} is defined as
\begin{equation}
H^{k+1}=\sigma(\tilde{D}^{-1} \tilde{A} H^kW^k) \label{conv2}.
\end{equation}
The above rule  can be obtained if $L=I_n-D^{-1}A$.

\section{The Method}
In this section, we first introduce the motivation of this study and then elaborate on the technical details of our proposed framework named CenGCN.
\begin{figure*}[t]
	\centering
	\includegraphics[width=7in]{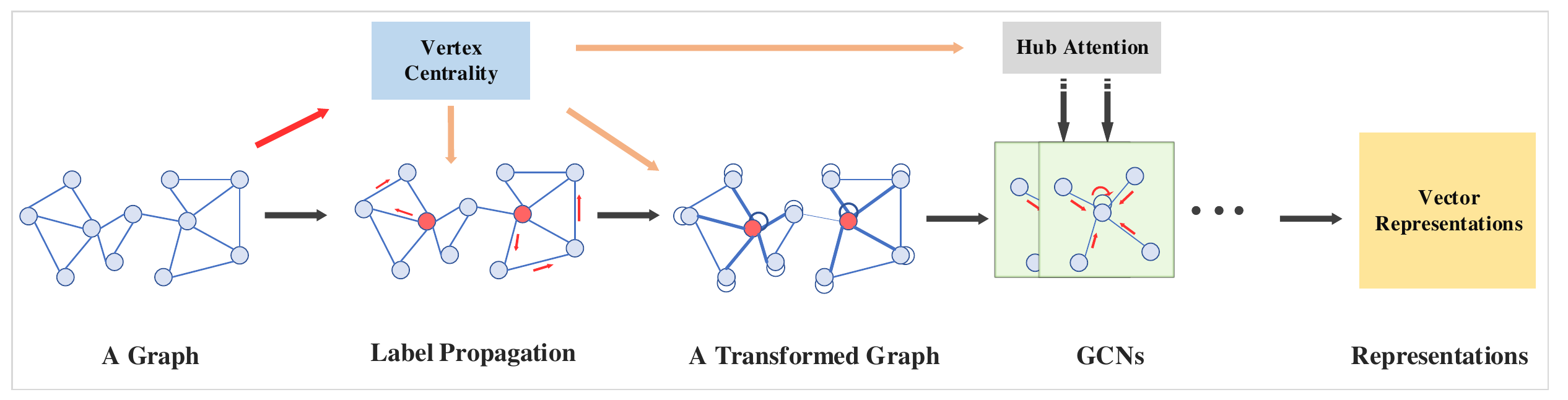}
	\caption{The framework of CenGCN consists of four steps: computing vertex centrality indices; label propagation using hub vertices, highlighted by red in the above feature; obtaining a transformed graph with self-connections; using multi-layer GCNs with hub attention to generate vector representations. Note that the dots represent repeated layers of GCNs.} \label{figure:framework}
\end{figure*}
\subsection{Motivation}
From Eq. \eqref{conv1} and Eq. \eqref{conv2}, it can be seen that GCNs leverage the immediate adjacency matrix to averagely aggregate information from neighbors and selves,  with the belief that all information from different sources is equally important. GCNs only consider inter-node connections when aggregating neighbor information, not vertex types and vertex information. In the real world, some networks have a scale-free property, such as social networks. In these networks, a vertex is more likely to be attracted by hub vertices with high centrality values than by ordinary vertices with low centrality values. Thus the information from hub vertices is more dominant. Existing GCNs, however, have not yet exploited such an important property. In this paper, we study how to define a generalized and transformed adjacency matrix that captures the influence of hub vertices on their neighbors, and how to use the transformed adjacency matrix to improve performance of GCNs.

Recently, Yan et al. \cite{yan2017graph,ghosh2014interplay} have proposed a transformed adjacency matrix defined as:
\begin{equation}
\tilde{A}=D(T-I)+BAB \label{trans_adj},
\end{equation}
where $B$, a biased diagonal matrix with each entry greater than zero, changes weights of all edges. $T$, a diagonal matrix where each entry $T_{ii}\geqslant 1$, adds a self-connection to each vertex. Because of various $B$ and $T$, the transformed adjacency matrix can support a wide variety of centrality indices and communities and is beneficial to capture underlying network characteristics. In Eq. \eqref{trans_adj}, the weights of self-connections are limited to multiples of their degrees. To generalize significantly the transformed adjacency matrix, we redefine it as:
\begin{equation}
\tilde{A}=T+BAB \label{redef_adj},
\end{equation}
where both $T_{ii}$ and $B_{ii}$ are greater than or equal to one. If we set $T=I$ and $B=I$, $\tilde{A}$ in Eq. \eqref{trans_adj} is equal to counterparts in Eq. \eqref{conv1} and \eqref{conv2}. Because of the generality and flexibility of $T$ and $B$, we can define manifold $\tilde{A}$, whereby GCNs employ multiple network characteristics. In this paper, we study how to incorporate vertex centrality into  $T$ and $B$, which is particularly important for scale-free networks.

A trivial solution is setting $T_{ii}$ and $B_{ii}$ to the centrality index of $v_i$.  However, one non-negligible issue is that hub vertices with high centrality are likely to attract dissimilar vertices, due to their high attractiveness. For instance, we may follow some persons merely because of their reputation on social media. Such dissimilarity will be strengthened if we use centrality indices to weight edges. To weaken the influence of hub vertices on dissimilar neighbors while transforming the adjacency matrix, we consider the underlying network structure. Numerous studies have indicated that the network structure implies the similarity between vertices \cite{perozzi2014deepwalk,grover2016node2vec,wang2016structural}.
A vertex pays more attention to hubs that show higher similarity with it in network structure. Therefore, instead of relying on the T and B matrices to transform the adjacency matrix, we design three functions to transform the adjacency matrix by combining the vertex centrality indices and the similarity between vertices, as compared to Eq. \eqref{redef_adj}. Finally, we define the transformed adjacency matrix as a combination of three functions:
\begin{equation}
\tilde{A}_{ij}=
\begin{cases}
f_C(v_i) & \text{if } i=j,\\
f_B(A_{ij},f_C(v_i), f_C(v_j), f_S(v_i,v_j)) & \text{if } i \neq j, \label{tran_a}
\end{cases}
\end{equation}
where $f_C: V \rightarrow \mathbb{R}$ tells us vertices' centrality indices, and $f_S: V\times V \rightarrow \mathbb{R}$ returns the similarity of two vertices in network structure. $f_B: \mathbb{R}^4 \rightarrow \mathbb{R}$ calculates a new weight for each pair of connected vertices. This definition of transforming adjacency matrix has two advantages:
\begin{itemize}
	\item Incorporating vertex centrality indices, vertices pay more attention to similar neighbors with higher centrality indices.
	\item Considering the underlying network structure, the influence of hub vertices on their dissimilar neighbors is reduced.
\end{itemize}

It is insufficient for GCNs to directly use the transformed adjacency matrix. This is because the transformation process ignores the influence of non-hub neighbors that are beneficial for the central vertex. We propose a hub attention mechanism, by which more information passes through edges whose two endpoints share many common hubs.

Next, we elaborate on how to define the three functions $f_C$, $f_B$, and $f_S$ and how to inject the hub attention into the layers of GCNs. Firstly, we describe the overall framework of CenGCN.

\subsection{Overall Framework}
The overall framework of CenGCN is shown in Fig. \ref{figure:framework}. Given a graph with scale-free property, we first compute the vertex centrality indices. Any vertex centrality measurement can be used here. Based on the computed vertex centrality, we can identify hub vertices and highlight them in red. After labeling these hub vertices, we propose a label propagation method to quantify the similarity between hubs and their neighbors. The proposed method is based on random walk that can reveal the similarity between vertices in network structure.
Using the quantified similarities and centrality indices, we transform the given graph to a new graph by increasing or decreasing the weight of each edge. Then self-connections are added to the new graph to force nodes to concentrate on their own characteristics. In the shown figure, the weight of each edge in the transformed graph is drawn proportionally as the thickness of the corresponding line. Then, the transformed graph is used to build multi-layer GCNs. Meanwhile, the hub attention mechanism is injected into each layer to enhance CenGCN.  The outputs of GCNs are vector representations of all vertices. These representations can be used for subsequent tasks, such as vertex classification.

Next, we detail each step of this framework.

\subsection{Vertex Centrality}
There exists various centrality measurements. In this study, we use degree centrality \cite{freeman1978centrality} and eigenvector centrality \cite{bonacich1972factoring}, but other centrality measurements are also applicable.

For degree centrality, we define $c_i=D_{ii}$. For eigenvector centrality, we first obtain $\lambda_{max}$, the maximum absolute eigenvalue of the adjacency matrix. After that, we compute the eigenvector $\vec{v}$ that corresponds to eigenvalue $\lambda_{max}$, according to Eq. \eqref{eigen_d}:
\begin{equation}
\lambda_{max}\vec{v}=A\vec{v} \label{eigen_d},
\end{equation}
where $\vec{v}_i$ is the centrality index of $v_i$. To ensure every value in $\vec{v}$ is greater than or equal to one, we define $\vec{v}'$=$abs(\vec{v})/min(abs(\vec{v}))$, where $abs(\vec{v})$ denotes changing each element in $\vec{v}$ to its absolute value, and $min(abs(\vec{v}))$ represents the minimum absolute value in $\vec{v}$. Finally, we define $c_i=\vec{v}'_i$.

After obtaining vertex centrality indices denoted by $c$,  we define those vertices with very high centrality as hub vertices. To be specific, vertices whose centrality indices are in the top $r\%$ $(0 < r < 100)$ are hubs. We denote the set of hub vertices by $N_h$ and use $r$ to denote the proportion of hub vertices.

The function $f_C(v_i)$ is defined as:
\begin{equation}
f_C(v_i)=
\begin{cases}
c_i & \text{if } v_i  \in N_h,\\
1 & \text{else}.
\end{cases}\label{fun_fc}
\end{equation}

We only consider the influence of hub vertices. Therefore, $f_C$ is defined to  maintain only the centrality indices of the hub vertices.

\subsection{Label Propagation}
To capture the similarity between hub vertices and their neighbors in the underlying network structure, we propose a label propagation method based on random walk. It is well established that random walk shows the similarity between vertices in the network structure \cite{Xia2019TETCIrandom}. As a result, it is widely used for community detection \cite{von2007tutorial,rosvall2008maps} and recommendations \cite{nikolakopoulos2019recwalk,Ren2021TETCI}. If two vertices are similar, there is a high probability to move from one vertex to the other vertex within a small number of hops. This proposed label propagation outputs the probabilities from hub vertices to their neighbors. These probabilities reflect their similarities in the network structure.

We first give each hub vertex a unique label and store labels in the matrix $L \in \mathbb{R}^{|N| \times |N_h|}$, where all elements are zero, but $L_{i, N_h \text{-} index(i)} = 1$ if $v_i \in N_h$. $N_h \text{-} index(i)$ denotes the index of hub vertex $i$ in the set of hub points $|N_h|$. We define $P=D^{-1}A$ as a probability transfer matrix, with $P_{ij}$ representing the probability of hopping immediately from $v_i$ to $v_j$.

After one propagation through Eq. \eqref{first_pro}, vertices obtain labels from hubs connected with them. Repeated propagations transmit labels of hubs to more vertices, the proportion of labels decreasing as the distance to hubs increases.
After $t$ ($t$ is set to 5 in this paper) propagations (Eq. \eqref{t_pro}), we denote the final label matrix by $L^t$. The $i_{th}$ row $L^t_i$ represent the probabilities of moving from $v_i$ to all hub vertices within 5 hops. The label score $L^t_{ij}$ is the specific probability of moving from $v_i$ to $v_j$, revealing how similar vertex $v_i$ and hub vertex $v_j$.
\begin{equation}
L^1=PL \label{first_pro},
\end{equation}
\begin{equation}
L^t=PL^{t-1}, \qquad t>1 \label{t_pro}.
\end{equation}

Given a vertex $v_i$, a hub vertex $v_j$, and score $L^t_{ij}$,  we cannot immediately decide whether $v_i$ and $v_j$ are accidentally connected or dissimilar, since $L^t_{ij}$ is greater than zero. If $e_{ij}$ connecting vertex $v_i$ and hub vertex $v_j$ is an accidental edge that is the dissimilar situation between vertex, the corresponding value $L^t_{ij}$ should be very small, but we need to decide the extent of the smallness. Here, we assume that a vertex should have a stronger relationship with hub vertices connected with it than with hub vertices not connected with it when no accidental links appear.  Given a vertex $v_i$ and its hub neighbors $N^h_i$,  we reward these connected hubs whose label scores are among the top  $|N^h_i|$ scores of $L^t_i$ and punish those linking hubs whose label scores are outside the top $|N^h_i|$ scores. Formally, we sort $L_i^t$ by decreasing order and define $Rank_i(j)$ as a function returning the rank of $L^t_{ij}$ in the sorted order. The $fs(v_i,v_j)$ is defined as:
\begin{equation}
f_S(v_i,v_j)=min(f_S'(v_i,v_j),f_S'(v_j,v_i)), \label{fun_fs}
\end{equation}
where
\begin{equation}
f_S'(v_i,v_j)=
\begin{cases}
1  &  \text{if }  v_j \notin N_h,\\
1 & \text{if }  v_j \in N_h \, \text{and} \, Rank_i(j)\leqslant |N_h^i|,\\
-1 & \text{if }  v_j \in N_h \, \text{and} \, Rank_i(j)> |N_h^i|.
\end{cases}
\end{equation}
$min(\cdot,\cdot)$ returns the minimum value, rendering $f_S(v_i,v_j)$ a symmetric function, i.e., $f_S(v_i,v_j)=f_S(v_j,v_i)$. If $f_S(v_i,v_j)=-1$, $v_i$ and $v_i$ are dissimilar, even though they link to each other. It is noted that if and only if $v_i \in N_h$ or $v_j \in N_h$, $f_S(v_i,v_j)$ has a chance of equaling to $-1$. The reasons are: (i) hub vertices are more likely to link to dissimilar neighbors than ordinary vertices; (ii) A low centrality index is unable to add large weights to neighbors.

\subsection{Graph Transformation}
Next, we define $f_B(A_{ij},f_C(v_i), f_C(v_j), f_S(v_i,v_j))$ as:
\begin{multline}
f_B(A_{ij},f_C(v_i), f_C(v_j), f_S(v_i,v_j))=\\
\begin{cases}
A_{ij}*f_C(v_i)^p* f_C(v_j)^p & \text{if  } f_S(v_i,v_j)=1,\\
A_{ij}*f_C(v_i)^{q}* f_C(v_j)^{q} & \text{if  } f_S(v_i,v_j)=-1,\\ \label{fun_fb}
\end{cases}
\end{multline}
where we use two hyper-parameters $p$ and $q$ $(p>0, q<0)$ to control the influence extent of vertex centrality indices.  If $f_S(v_i,v_j)=1$, centrality indices are used to weight $A_{ij}$. Otherwise, we reduce the weight of $A_{ij}$ using centrality indices.

After defining $f_C$, $f_S$ and $f_B$, we can obtain the transformed adjacency matrix $\tilde{A}$. The transformed matrix not only incorporates vertex centrality indices, but also considers the underlying structure. We summarize this process of graph transformation in Algorithm \ref{alg_1}.

\begin{algorithm}[htbp]
	\caption{Graph Transformation}\label{alg_1}
	\begin{algorithmic}[1]
		\REQUIRE A graph $G=(V,E)$, the adjacency matrix $A$, hub rate $r$, propagation number $T$, a centrality measurement, and hyper-parameters $p$ and $q$.
		\ENSURE A transformed adjacency matrix $\tilde{A}$.
		\STATE Compute centrality index $c_i$ for $v_i \in V$
		\STATE Obtain hub vertices $N^h$ whose centrality indices are in top $r\%$
		\STATE Define $f_C$ according to Eq. \eqref{fun_fc}
		\STATE Define label matrix $L$ and probability matrix $P$
		\STATE $L^1=PL$
		\FOR {$t=2$ to $T$}
		\STATE $L^t=PL^{t-1}$
		\ENDFOR
		\STATE Define $f_S$ according to Eq. \eqref{fun_fs}
		\STATE Define $f_B$ according to Eq. \eqref{fun_fb}
		\STATE Obtain $\tilde{A}$ according to Eq. \eqref{tran_a}
	\end{algorithmic}
\end{algorithm}

\subsection{Hub Attention}
We define the convolution operation of GCNs at $k_{th}$ layer as
\begin{equation}
H^k=\sigma(\tilde{D}^{-1}\tilde{A}H^{k-1}W^{k-1}) \label{pro_rule},
\end{equation}
where $\sigma$ an activation functions, set to $tanh$ in this study. For vertex $v_i$, this operation also can be written as:
\begin{equation}
H^k_i=\sigma(\frac{1}{\tilde{D}_{ii}}\sum_{v_j\in N_i\cup\{v_i\}}\tilde{A}_{ij}H^{k-1}_jW^{k-1}). \label{pro_rule_i}
\end{equation}

From the above equation, we can see that if $v_i$ is connected to a hub vertex with an extremely high centrality index, the information flowing into $v_i$ is almost totally from this hub. The information from non-hub neighbors plays an important role in the decision of the central vertex, such as deciding which class it belongs to. We propose a hub attention mechanism which assigns new weights to non-hub neighbors by consideration of common information from hub vertices. After the convolution of the transformed graph, non-hubs with many shared hub vertices will have similar features between them, and the attention mechanism will assign large weights between vertices with similar features. Therefore, a large weight is assigned to two connected non-hub vertices that share significant hub information. We define $\tilde{N}_i^{h}=N_i-N^h_i$ as the set of non-hub neighbors of $v_i$.
At $k_{th}$ layer, the weight between $v_i$ and $v_j$ is defined as:
\begin{equation}
a_{ij}=\frac{exp(H^k_i \cdot H^k_j)}{\sum_{v_l \in \tilde{N}_i^{h}\cup\{v_i\} }exp(H^k_i \cdot H^k_l)},
\end{equation}
where $\cdot$ represents the dot product of two vectors. Based on the hub attention, a new convolution is defined as:
\begin{equation}
\tilde{H}^k_i=\sigma(\sum_{v_j\in \tilde{N}_i^{h}\cup\{v_i\}}a_{ij}H_j^k). \label{att_pro}
\end{equation}
The resulting $\tilde{H}^k_i$ is concatenated with $H^k_i$ to enhance GCNs. Finally, $H^k_i$ is computed anew as $H^k_i=H^k_i||\tilde{H}^k_i$, where $||$ represents the concatenation of two vectors. The hub attention mechanism is significantly different with GATs \cite{velickovic2018graph}. The differences include:
\begin{enumerate}
	\item GATs learn individual representations. The representations learned by the hub attention is used as vital complements to GCNs.
	\item GATs define many matrices to learn attention weights and require multi-head attention to maintain stability. The hub attention does not need these and thus it is faster to compute attention weights.
	\item GATs learn attention weights using features of the previous layer. The hub attention learns them using features of the current layer.
\end{enumerate}

\subsection{Optimization}
Suppose that CenGCN uses $K$-layer GCNs, the final vector representation is $Z=H^K$. The learned $Z$ can be used for several network-based tasks. To train CenGCN, we consider both semi-supervised learning and unsupervised learning in case there is no supervision information available.

\textbf{Semi-supervised Learning.} Let $Y$ denote ground-truth vertex class and $\mathcal{Y}_L$ denotes the set of node indices that have class information. We use  cross-entropy as the loss function:
\begin{equation}
\mathcal{L}=-\sum_{l\in \mathcal{Y}_L}\sum_{f=1}^{F}Y_{lf}\ln Z_{lf}, \label{supervised_loss}
\end{equation}
where $Y_{lf}=1$ indicates that $v_l$ belongs to class $f$, while $Y_{lf}=0$ indicates otherwise.

\textbf{Unsupervised Learning.} For unsupervised learning, the loss function is defined by reconstructing edges in the original graph:
\begin{equation}
\mathcal{L}=||(sigmoid(ZZ')-A)\otimes \hat{A}||^2_F, \label{unsupervised_loss}
\end{equation}
where $Z'$ is the transpose of $Z$. $||\cdot||^2_F$ is the squared Frobenius norm. $\otimes$ represents element-wise matrix multiplication. In the adjacency matrix $A$, zero elements outnumber non-zero elements, particularly for sparse networks. As a result, the unsupervised learning is prone to reconstruct zero elements of $A$. In the above equation, we thus define a matrix $\hat{A}$ to attach higher weights to non-zero elements of $A$. Specifically, $\hat{A}_{ij}=\rho$ $(\rho>1)$ if $A_{ij}=1$, else $\hat{A}_{ij}=1$. Here, we set $\rho$ to 100.

The final loss function is defined as:
\begin{equation}
\mathcal{L}_{loss}=\mathcal{L}+\alpha\mathcal{L}_{reg} \label{loss},
\end{equation}
where $\mathcal{L}_{reg}$ is the regularization loss of all learned weights, defined as $\sum_{k=0}^{K-1}||W^k||^2_F$. $\alpha$ is the hyper-parameter set to  $5\times 10^{-4}$.
To minimize the loss $\mathcal{L}_{loss}$ and update the parameters of CenGCN, we employ $Adam$ \cite{kingma2015adam,wang2020model} and $Dropout$ \cite{srivastava2014dropout,liu2019shifu2} with $keep\_pro=0.5$.

We summarize the overall framework of CenGCN in Algorithm \ref{alg_2}.
\begin{algorithm}[htbp]
	\caption{The Framework of CenGCN}\label{alg_2}
	\begin{algorithmic}[1]
		\REQUIRE A graph $G=(V,E)$, the adjacency matrix $A$, the feature matrix $X$, hub rate $r$, propagation number $T$, a centrality measurement, number of layers $K$, hyper-parameters $p$, $q$ and $\alpha$, learning rate $\theta$, class information $\mathcal{Y}_L$, and a convergence condition.
		\ENSURE Well trained  CenGCN.
		\STATE Obtain the transformed adjacency matrix $\tilde{A}$ and $N^h$ using Algorithm \ref{alg_1}
		\STATE Initialize all weight parameters $\{W^k\}_{0\leq k \leq K-1}$
		\WHILE {The convergence condition is not satisfied:}
		\STATE $H^0$=$X$
		\FOR {$k=1$ to $K$}
		\STATE Compute $H^k$ according to \eqref{pro_rule}
		\STATE Compute$\tilde{H}^k$ according to \eqref{att_pro}
		\STATE $H^k=H^k||\tilde{H}^k$
		\ENDFOR
		\STATE $Z=H^K$
		\STATE Compute $\mathcal{L}_{reg}=\sum_{k=0}^{K-1}||W^k||^2_F$
		\IF{ class information is available:}
		\STATE Compute $\mathcal{L}$ according to \eqref{supervised_loss}
		\ELSE
		\STATE Compute $\mathcal{L}$ according to \eqref{unsupervised_loss}
		\ENDIF
		\STATE Compute $\mathcal{L}_{loss}$ according to \eqref{loss}
		\STATE Minimize $\mathcal{L}_{loss}$ by $Adam$ with learning rate $\theta$
		\ENDWHILE
	\end{algorithmic}
\end{algorithm}

\subsection{Computational Complexity}
The first step of the CenGCN framework is calculating centrality indices. The time complexity at this step is  $\Theta(n)$ when degree centrality is used and $\Theta(n^2)$ when eigenvector centrality is used. The next step is label propagation and can be finished in $\Theta(Tn^2)$ since here we only calculate $T$ dot productions of a vector and a matrix. At last, the time complexity in GCN layers is $\Theta(Ln^2)$, where $L$ here is the number of layers. Overall, the computational complexity of the CenGCN framework is $\Theta((T+L)n^2)$.

\section{Experiments}
In this section, we compare our proposed framework CenGCN with several baselines by running four experiments: vertex classification, link prediction, vertex cluster, and network visualization. The results of parameter sensitivities are presented at the end of the section. We have implemented the CenGCN in Python 3.6 with Tensorflow1.15.
\subsection{Datasets}
We use five datasets, which are introduced in \cite{leskovec2012learning}, \cite{feng2018representation} and \cite{yang2015defining}.
Their statistics are summarized in Table \ref{table:datasets}.
\begin{itemize}
	\item Facebook: The data was collected from survey participants using the Facebook app. Vertices represent users, and edges represent friendship.
	\item Twitter:  The data was crawled from public sources. Vertices indicate users, and edges denote following relationships.
	\item Gplus: The data was collected from Google+. Vertices indicate users, and edges denote following relationships.
	\item Youtube:  The data was collected from a video-sharing website that includes a social network. Users are denoted by vertices, and edges denote friendship.
	\item LiveJournal: The data was collected from a free on-line blogging community, where users declare friendship with each other. Vertices represent users, and edges represent  friendship.
\end{itemize}
\par The above datasets are scale-free networks. Most of the connections are concentrated in a few centers. In the semi-supervised learning task (vertex classification), for all five datasets we use vertex classes as labels.
\begin{table}[htbp]
	\centering
	\caption{ Statistics of datasets. '---' means no data available.} \label{table:datasets}
	\begin{tabular}{c|c|c|c|c}
		\toprule
		datasets &  Vertices & Edges & Features & Classes\\
		\midrule
		Facebook & 3944 & 87870 & 1385 & 8\\
		Twitter & 1533 & 38323 & 10353 & 8\\
		Gplus & 5331 & 351726 & 1988 & 5\\
		Youtube & 4684 & 19443 & ---& 8\\
		LiveJournal & 3009 & 44599 & ---& 4\\
		\bottomrule
	\end{tabular}
\end{table}
\subsection{Comparison Algorithms}
Based on two different centrality measurements, we define two variants of CenGCN:
\begin{itemize}
	\item CenGCN\_D: Utilizes degree centrality within the overall proposed framework.
	\item CenGCN\_E: Utilizes eigenvector centrality within the overall proposed framework.
\end{itemize}

In addition, we define the following variants from CenGCN\_D and CenGCN\_E as complements to demonstrate the efficacy and necessity of each part of CenGCN.
\begin{itemize}
	\item  CenGCN\_TD and CenGCN\_TE: Uses only the transformed adjacency matrix, without the hub attention mechanism.
	\item CenGCN\_AD, CenGCN\_AE: Uses only the hub attention mechanism, without the transformed adjacency matrix.
	\item CenGCN\_WD, CenGCN\_WE: Uses the centrality indices of  hub vertices to increase edge weights by setting $p$ to $q$.
	\item CenGCN\_ID, CenGCN\_IE: Uses the centrality indices of  hub vertices to decrease edge weights by setting $p$ to $q$.
\end{itemize}

To verify the efficiency of CenGCN, we conduct experiments against the following baselines:
\begin{itemize}
	\item GCN\_Cheby \cite{defferrard2016convolutional}: Uses fast localized convolutional filters on graphs using Chebyshev expansion.
	\item GCNs \cite{kipf2017semi}: Uses a layer-wise convolutional operation that encodes both local graph structure and vertex features.
	\item GATs \cite{velickovic2018graph}: Leverages masked self-attentional layers to specify different weights to different vertices in a neighborhood.
	\item DGI \cite{velickovic2019deep}: Learns vertex representations in an unsupervised manner, by relying on maximizing mutual information between patch representations and corresponding high-level summaries of graphs.
	\item H-GCN \cite{hu2019hierarchical}: Repeatedly aggregates structurally similar vertices to hyper-vertices and then refines the coarsened graph to the original to restore the representation for each vertice.
	\item DPSW \cite{feng2018representation}: Punishes the proximity between high-degree vertices using scale-free property preserving network embedding algorithm. DPSW represents the best model drawing upon DP-Spectral and DP-Walker.
\end{itemize}

\subsection{Experimental Setup}
We consider the four different network tasks:
\begin{itemize}
	\item Vertex classification: This is a semi-supervised learning task. Classes of vertices are the ground truth. 10\% vertices with class information are used as training examples, and 10\% vertices with class information are validation examples. The remaining vertices are test examples. The learning rate $\theta$ is set to 0.01, and the iteration number is set to 1000. The best parameters on validation examples are saved and then used for test examples. Accuracy is used as the evaluation metric.
	\item Link prediction: This is an unsupervised learning task. We first randomly hide 50\% edges as positive examples and randomly select 50\% non-existent edges as negative examples. The remaining graph is used to train. According to the paper \cite{grover2016node2vec}, the Hadamard operator of two vertices is a good representation for their edge. Thus, we construct edge representations by this operator. Logistic regression is used for binary classification. The learning rate $\theta$ is set to 0.01. We stop the training when the loss $\mathcal{L}_{loss}$ remains stable or the iteration number is over 150. AUC (Area Under the Curve) is used as the evaluation metric.
	\item Vertex clustering: This is an unsupervised learning task. The representation $Z$ serves as the input features of K-means, a clustering method. Classes of vertices are the ground truth. Normalized Mutual Information (NMI) \cite{estevez2009normalized} is used as the evaluation metric. The learning rate $\theta$ is set to 0.001, a smaller rate 0.00001 on Twitter. We stop the training when the loss $\mathcal{L}_{loss}$ remains stable or the iteration number is over 150.
	\item Network visualization: The representation $Z$ obtained in vertex clustering is used here for network visualization. We feed $Z$ into the standard t-SNE tool \cite{maaten2008visualizing} to lay out the network and mask vertices of the same class with the same color. The network is visualized in a 2-dimensional space.
\end{itemize}

In semi-supervised learning, we employ a two-layer GCN with a 16-unit hidden layer for all variants of CenGCN. In unsupervised learning, we employ a two-layer GCN with a 512-unit hidden layer and a 128-unit output layer for all variants of CenGCN. The settings and sensitivities of parameters  $p$, $q$, and $r$, as well as the number of layers, are presented in \textit{Parameter Sensitivity}. The parameters of the baselines are set in accordance with the original papers.

\begin{table}[htbp]
	\setlength\tabcolsep{3pt}
	\centering
	\caption{ The accuracy of vertex classification. The best performance is boldfaced.} \label{table:classification}
	\begin{tabular}{c|c|c|c|c|c}
		\toprule
		Algorithm &  Facebook & Twitter & Gplus & Youtube & LiveJournal\\
		\midrule
		GCN\_Cheby & 0.915 & 0.972 & 0.787 & 0.812 & 0.810 \\
		GCNs & 0.914 & 0.954 & 0.716 & 0.889 & 0.901 \\
		GATs & 0.970 & 0.967 & 0.732 & 0.827 & 0.892 \\
		DGI & 0.936 & 0.954 & 0.771 & 0.227 & 0.592 \\
		H-GCN & 0.982 & 0.943 & 0.914 & 0.915 & 0.888 \\
		DPSW & 0.892 & 0.789 & 0.922 & 0.892 & 0.872 \\
		\midrule
		CenGCN\_D & \textbf{0.992} & \textbf{0.987} & \textbf{0.949} & \textbf{0.920} & \textbf{0.912} \\
		CenGCN\_TD & 0.970 & 0.982 & 0.943 & 0.914 & 0.897 \\
		CenGCN\_AD & 0.970 & 0.969 & 0.933 & 0.915 & 0.910 \\
		CenGCN\_WD & 0.832 & 0.965 & 0.941 & 0.904 & 0.903 \\
		CenGCN\_ID & 0.888 & 0.967 & 0.861 & 0.893 & 0.893 \\
		\midrule
		CenGCN\_E & \textbf{0.992} & \textbf{0.987} & 0.936 & 0.919 & 0.903 \\
		CenGCN\_TE & 0.912 & 0.905 & 0.717 & 0.873 & 0.894 \\
		CenGCN\_AE & 0.916 & 0.930 & 0.742 & 0.866 & 0.900 \\
		CenGCN\_WE & 0.932 & 0.973 & 0.717 & 0.892 & 0.895 \\
		CenGCN\_IE & 0.912 & 0.971 & 0.870 & 0.902 & 0.877 \\
		\bottomrule
	\end{tabular}
\end{table}

\subsection{Vertex classification}
The task of vertex classification is discovering classes of those vertices that have no class information. We first verify the efficacies of CenGCN and baselines through this task in this experiment. Table \ref{table:classification} shows the accuracies of CenGCN's variants and baselines on vertex classification. The best performance is boldfaced. From the table, we can see that on the five networks, CenGCN\_D always achieves the best performance and CenGCN\_E outperforms all baselines. These results demonstrate the significant superiority of CenGCN and the necessity to incorporate vertex centrality indices into GCNs. Besides, the following findings are also striking:
\begin{itemize}
	\item  On two networks, Facebook and Twitter, CenGCN\_D and CenGCN\_W have the same performance. But on the other three networks, CenGCN\_D outperforms CenGCN\_W. Overall, CenGCN\_D performs better CenGCN\_E, owing to the scale-free property based on vertex degrees. The finding that CenGCN\_E outperforms all baselines gives us motivation to explore more centrality measurements.
	\item  This table shows that CenGCN\_D and CenGCN\_E consistently outperform DPSW on the five networks, though they are proposed for scale-free networks.  The difference in performance can likely be attributable to the fact that a hub vertex can link to both similar and dissimilar vertices, while DPSW assumes that a vertex with a higher degree is more dissimilar to its neighbors.
	\item Compared with standard GCNs, CenGCN\_D and CenGCN\_E achieve great performance. On Gplus, CenGCN\_D achieves an improvement of 23.6\%. Compared with state-of-the-art GCN-based variants, CenGCN\_D outperforms GATs by 22.0\% on Gplus, outperforms DGI by 70.1\% on Youtube, and outperforms H-GCN by 4.5\% on Twitter. These significant improvements indicate the necessity for GCNs to utilise vertex centrality.
	\item  CenGCN\_D outperforms the other four variants of CenGCN that use degree centrality; CenGCN\_E outperforms the other four variants of CenGCN that use eigenvector centrality. These results indicate that the transformed graph needs to be combined with the hub attention mechanism, and we need to consider both the increase and decrease of edge weights.
\end{itemize}

\begin{table}[htbp]
	\setlength\tabcolsep{3pt}
	\centering
	\caption{ The AUC score of link prediction. The best performance is boldfaced.} \label{table:link_pre}
	\begin{tabular}{c|c|c|c|c|c}
		\toprule
		Algorithm &  Facebook & Twitter & Gplus & Youtube & LiveJournal\\
		\midrule
		GCN\_Cheby & 0.672 & 0.842 & 0.725 & 0.676 & 0.759 \\
		GCNs & 0.809 & 0.729 & 0.711 & 0.578 & 0.711 \\
		GATs & 0.633 & 0.852 & 0.558 & 0.685 & 0.757 \\
		DGI & 0.723 & 0.862 & 0.678 & 0.613 & 0.621 \\
		H-GCN & 0.708 & 0.564 & 0.601 & 0.656 & 0.739 \\
		DPSW & 0.767 & 0.581 & 0.797 & 0.714 & 0.753 \\
		\midrule
		CenGCN\_D & \textbf{0.892} & \textbf{0.873} & \textbf{0.801} & \textbf{0.731} & 0.848 \\
		CenGCN\_TD & 0.854 & 0.857 & 0.787 & 0.718 & 0.850 \\
		CenGCN\_AD & 0.885 & 0.855 & 0.775 & 0.713 & 0.837 \\
		CenGCN\_WD & 0.884 & 0.850 & 0.796 & 0.728 & 0.831 \\
		CenGCN\_ID & 0.882 & 0.847 & 0.699 & 0.713 & 0.828 \\
		\midrule
		CenGCN\_E & 0.891 & 0.871 & 0.769 & 0.727 & \textbf{0.853} \\
		CenGCN\_TE & 0.887 & 0.858 & 0.753 & 0.715 & 0.850 \\
		CenGCN\_AE & 0.868 & 0.856 & 0.746 & 0.756 & 0.841 \\
		CenGCN\_WE & 0.808 & 0.861 & 0.742 & 0.698 & 0.848 \\
		CenGCN\_IE & 0.840 & 0.856 & 0.776 & 0.681 & 0.842 \\
		\bottomrule
	\end{tabular}
\end{table}
\subsection{Link Prediction}
Link prediction aims at predicting whether two vertices that are not connected are potentially connected. In this experiment, we concentrate on the link prediction task and compare the performance of CenGCN and baselines. Table \ref{table:link_pre} shows their AUC scores on link prediction. We report the best performance by boldface. From these AUC scores in the table, we can see that CenGCN performs extremely well.
Among the five maximum scores, CenGCN\_D achieves four and CenGCN\_E achieves one. These results suggest the vertex centrality is a powerful indicator of link prediction. More noticeable findings are summarized as follows:
\begin{itemize}
	\item As a whole, CenGCN\_D performs better than CenGCN\_E. On Livejournal, CenGCN\_E outperforms CenGCN\_D only by 0.3\%. We also notice that the AUC gap between them is small, except for Gplus where CenGCN\_D outperforms CenGCN\_E by up to 5.2\%. Although the vertex degree is an intuitive centrality measurement for scale-free networks, the results of eigenvector centrality compare very favourably to these of degree centrality.
	\item CenGCN\_D outperforms all baselines, with the largest improvement of 9.0\% on Facebook. CenGCN\_E outperforms all baselines in the vast majority of cases, with the largest improvement of 8.9\% on Facebook. It is an unanticipated finding that DPSW performs better CenGCN\_E on Gplus. But on the other four networks, CenGCN\_E performs better than DBSW, particularly on Twitter.
	\item In most cases, CenGCN\_D is the best among all variants with degree centrality, and CenGCN\_E is the best among all variants with  eigenvector centrality. Contrary to expectations, CenGCN\_IE outperforms CenGCN\_E on Gplus and CenGCN\_TD outperforms CenGCN\_D on Livejournal. Overall, CenGCN\_D or CenGCN\_E performs best only when all designed parts are used.
\end{itemize}
\begin{table}[htbp]
	\setlength\tabcolsep{3pt}
	\centering
	\caption{ The NMI of vertex clustering. The best performance is boldfaced.} \label{table:clustering}
	\begin{tabular}{c|c|c|c|c|c}
		\toprule
		Algorithm &  Facebook & Twitter & Gplus & Youtube & LiveJournal\\
		\midrule
		GCN\_Cheby & 0.540 & 0.729 & 0.389 & 0.379 & 0.636\\
		GCNs & 0.561 & 0.656 & 0.288 & 0.300 & 0.657\\
		GATs & 0.577 & 0.756 & 0.440 & 0.552 & 0.685\\
		DGI & 0.639 & 0.799 & 0.341 & 0.056 & 0.143\\
		H-GCN & 0.683 & 0.714 & 0.317 & 0.424 & 0.686\\
		DPSW & 0.342 & 0.506 & 0.211 & 0.642 & 0.512\\
		\midrule
		CenGCN\_D & \textbf{0.885} & \textbf{0.911} & \textbf{0.717} & \textbf{0.700} & \textbf{0.783} \\
		CenGCN\_TD & 0.860 & 0.881 & 0.613 & 0.668 & 0.767\\
		CenGCN\_AD & 0.838 & 0.909 & 0.620 & 0.657 & 0.764\\
		CenGCN\_WD & 0.784 & 0.808 & 0.326 & 0.598 & 0.738  \\
		CenGCN\_ID & 0.581 & 0.751 & 0.417 & 0.207 & 0.685 \\
		\midrule
		CenGCN\_E & 0.752 & 0.907 & 0.495 & 0.670 & 0.761 \\
		CenGCN\_TE & 0.672 & 0.757 & 0.441 & 0.619 & 0.774\\
		CenGCN\_AE & 0.732 & 0.905 & 0.468 & 0.667 & 0.739\\
		CenGCN\_WE & 0.709 & 0.875 & 0.387 & 0.654 & 0.750  \\
		CenGCN\_IE & 0.738 & 0.881 & 0.240 & 0.623 & 0.721 \\
		\bottomrule
	\end{tabular}
\end{table}

\begin{figure*}[t]
	\centering
	\subfloat[GCN\_Cheby]{\includegraphics[scale=0.22]{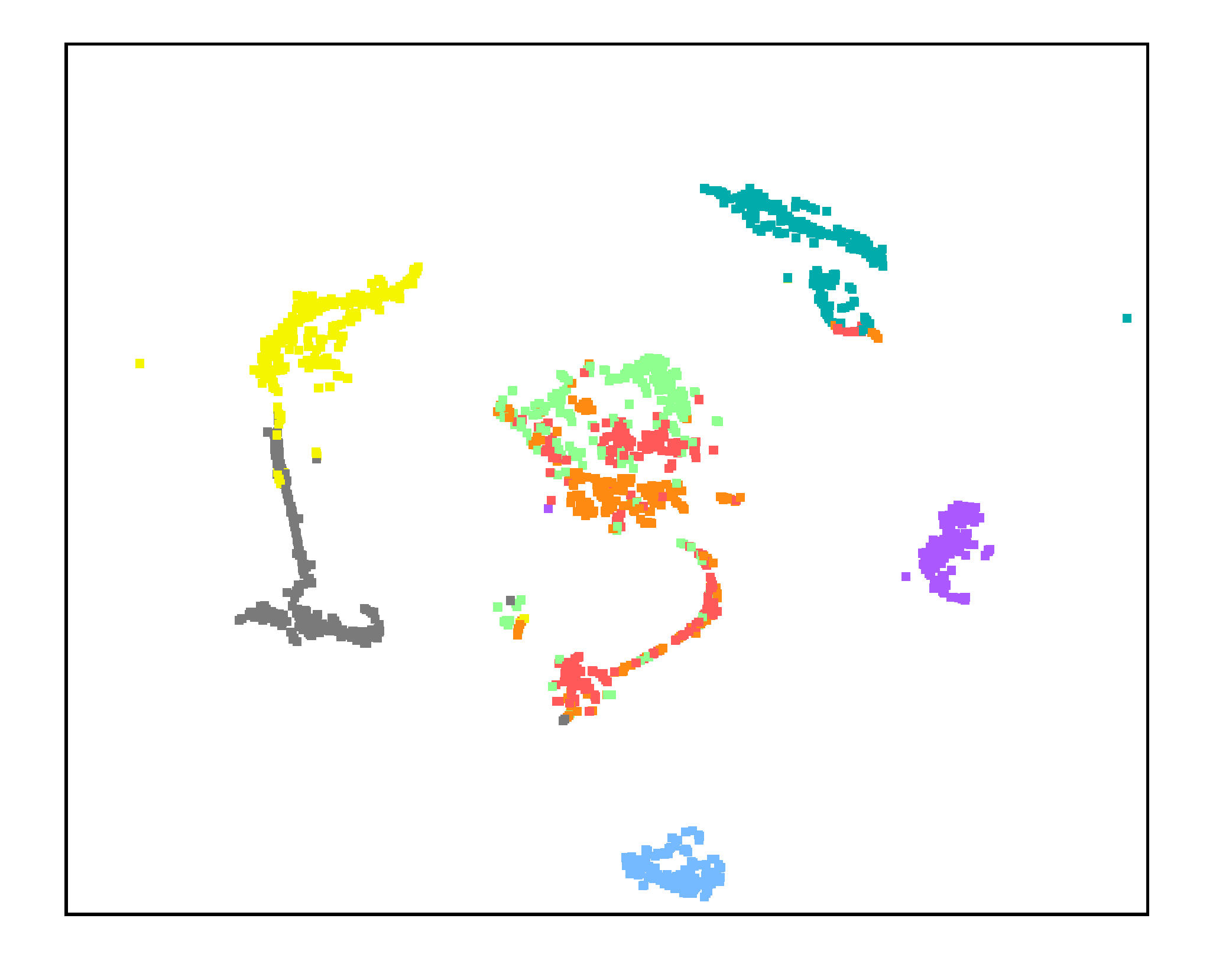}}
	\subfloat[GCNs]{\includegraphics[scale=0.22]{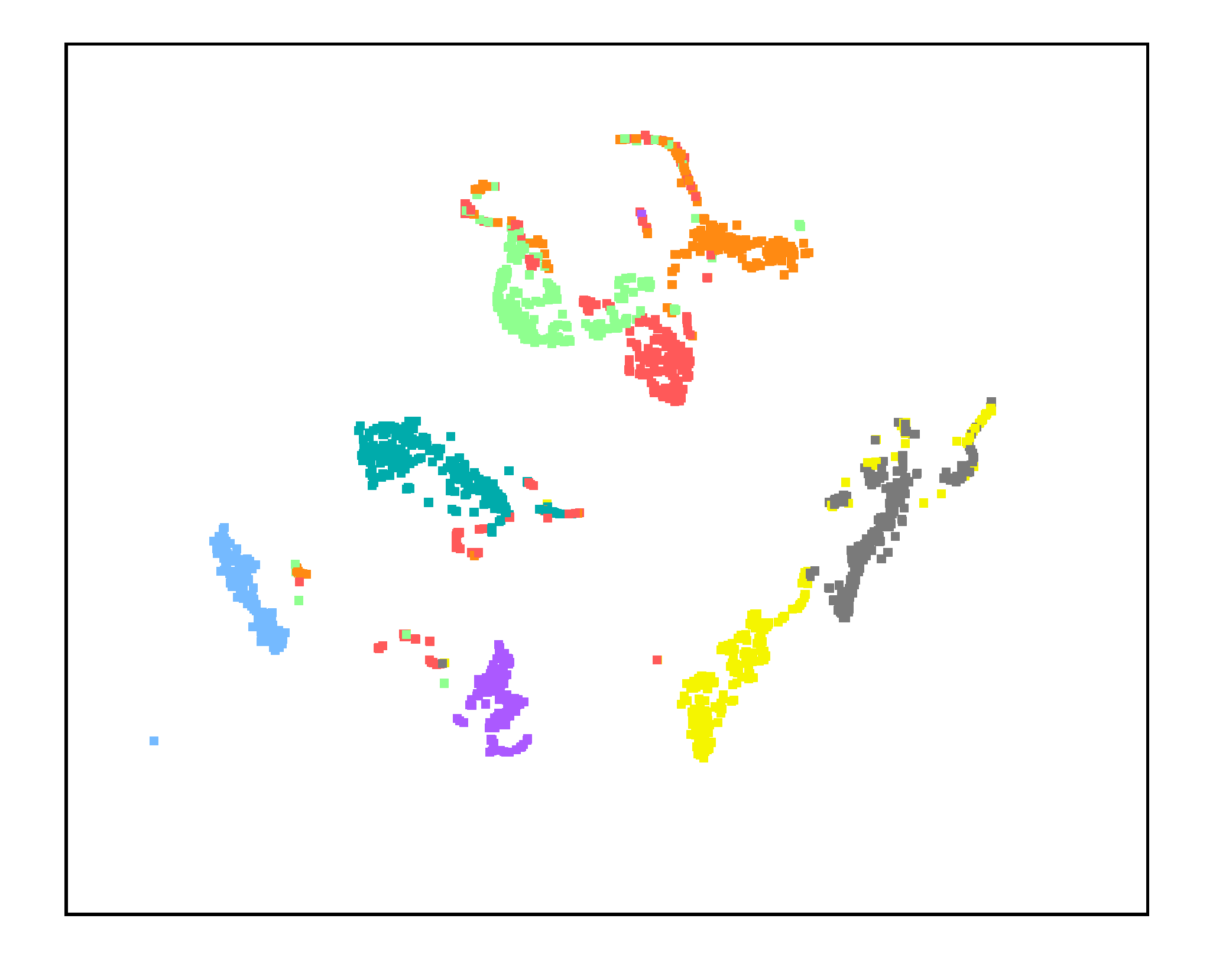}}
	\subfloat[GATs]{\includegraphics[scale=0.22]{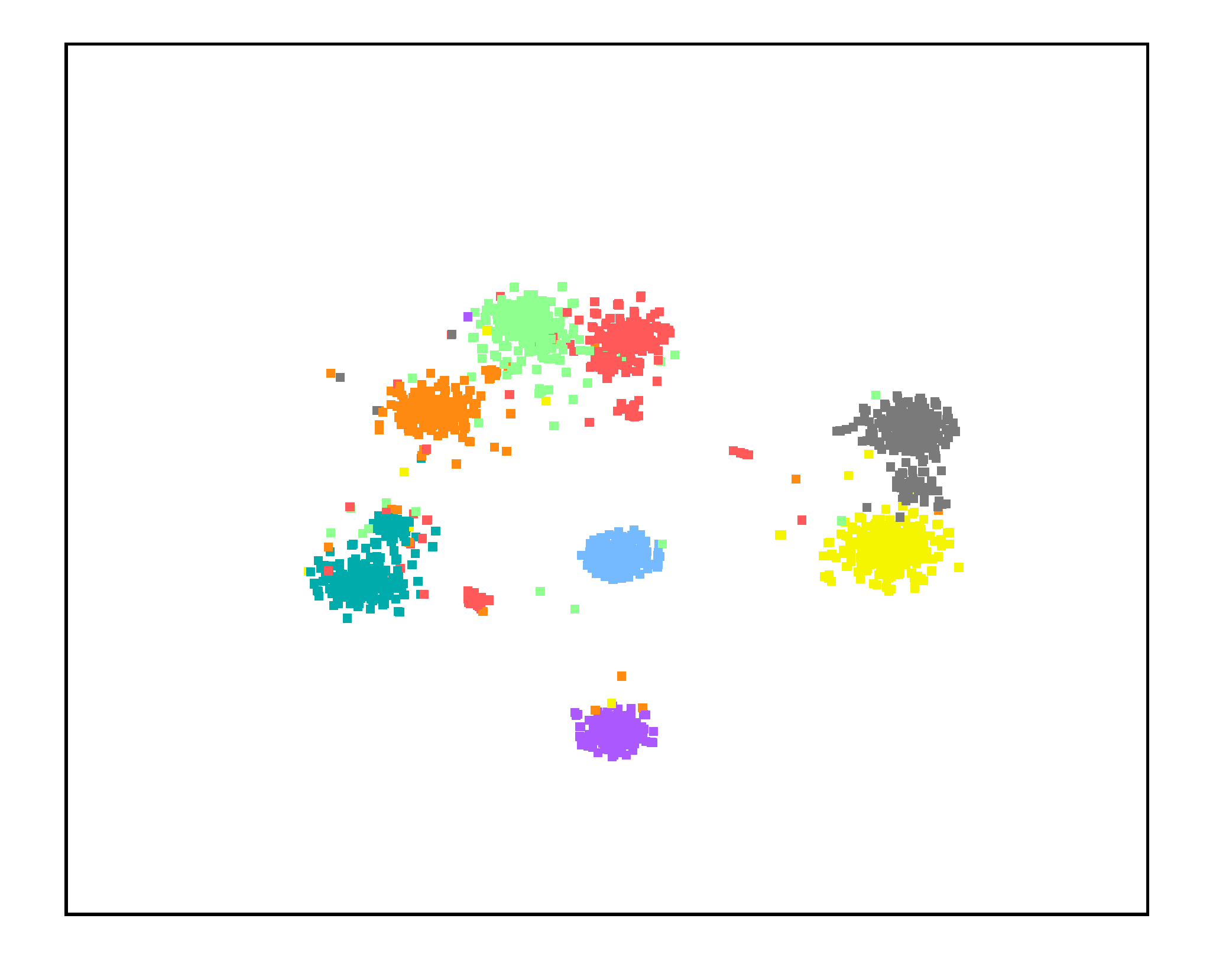}}
	\subfloat[DGI]{\includegraphics[scale=0.22]{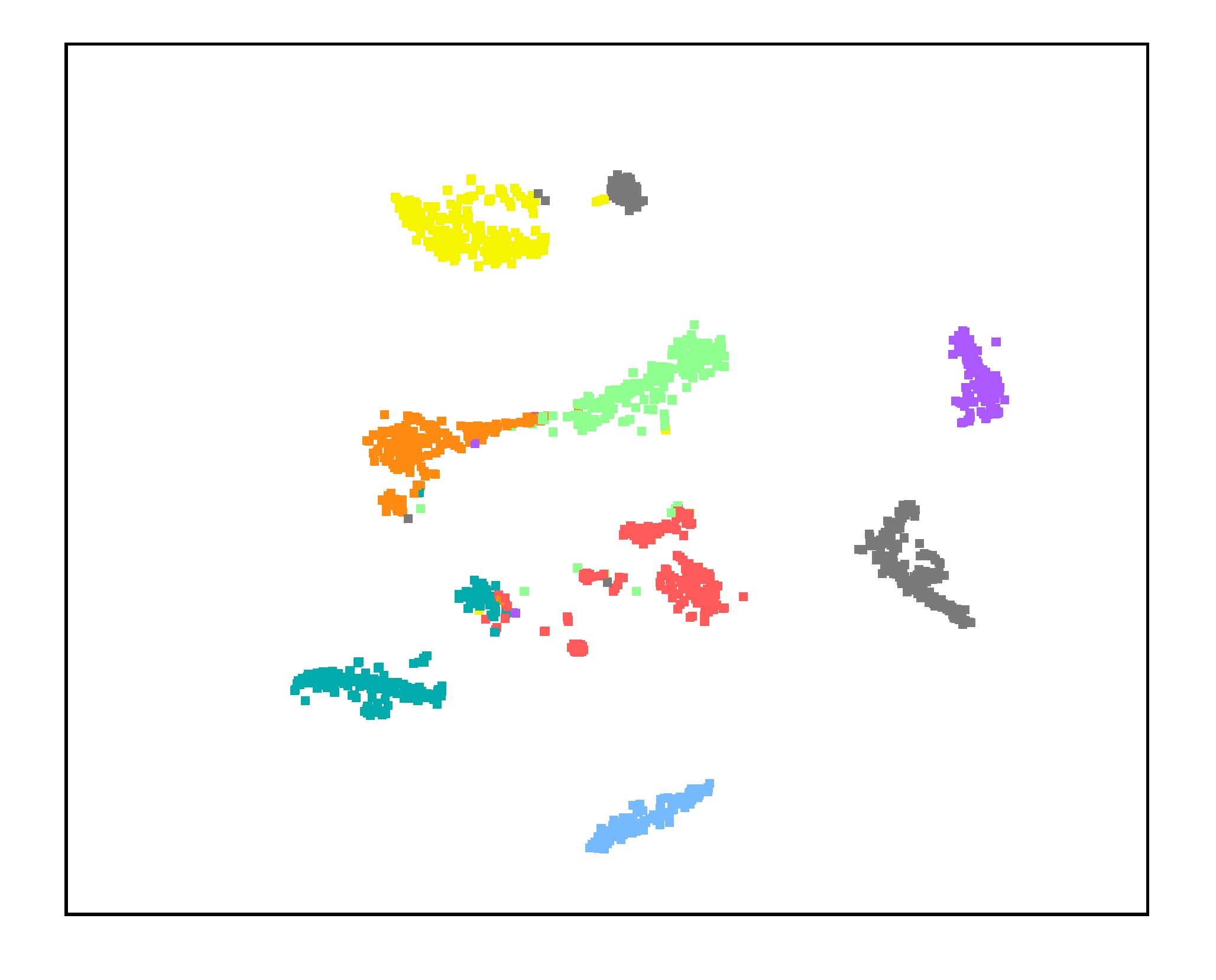}}
	\hfil
	\subfloat[H-GCN]{\includegraphics[scale=0.22]{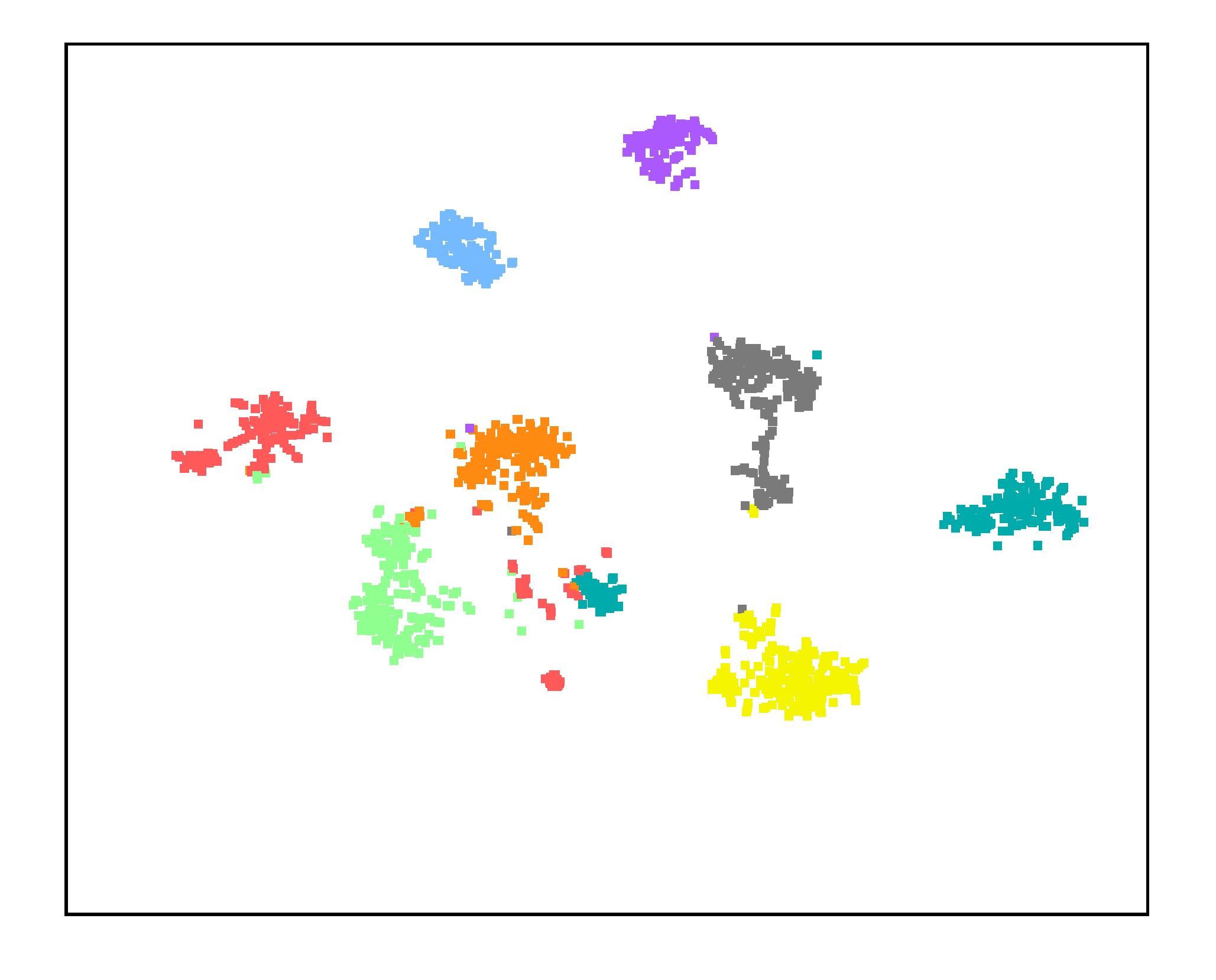}}
	\subfloat[DPSW]{\includegraphics[scale=0.22]{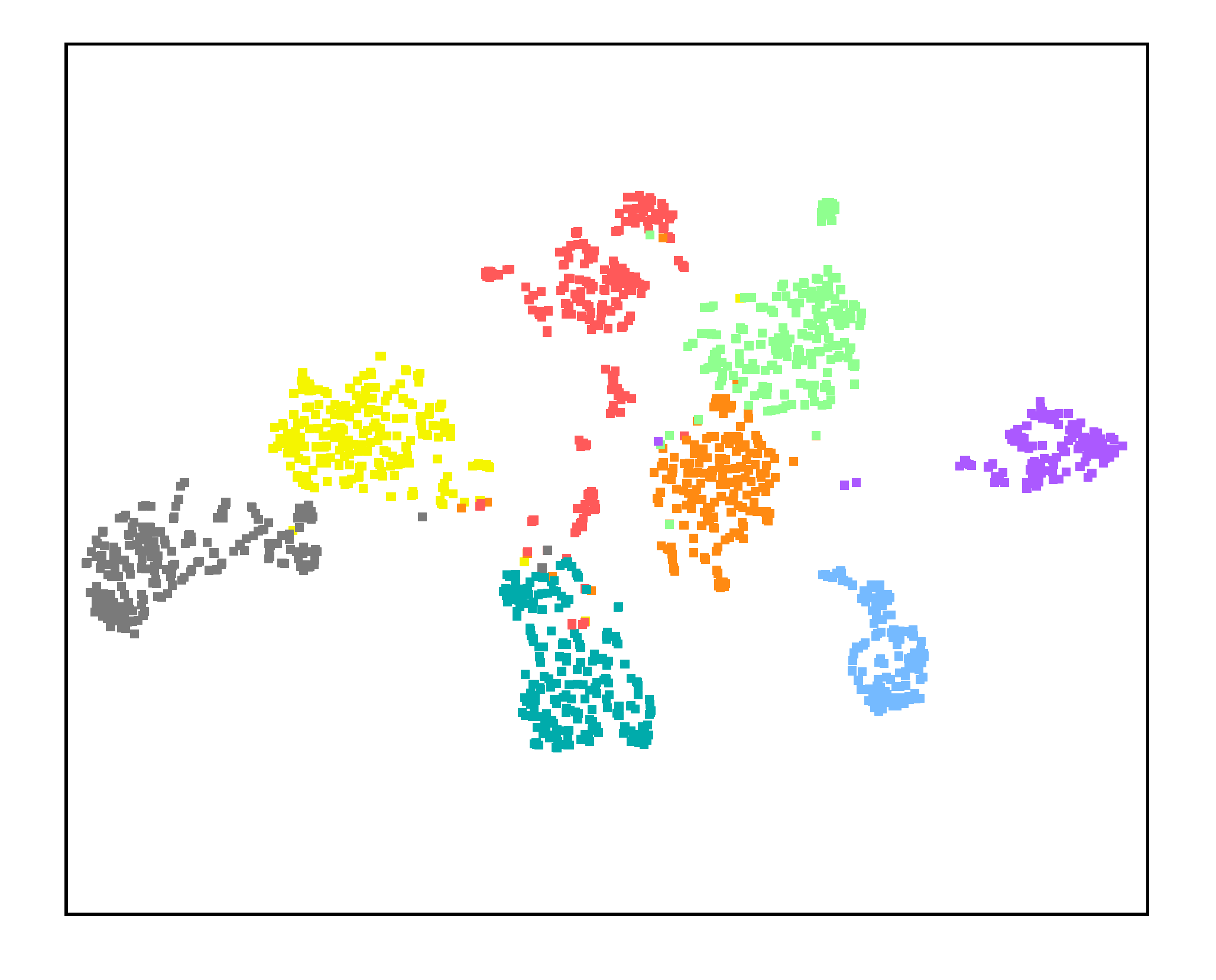}}
	\subfloat[CenGCN\_D]{\includegraphics[scale=0.22]{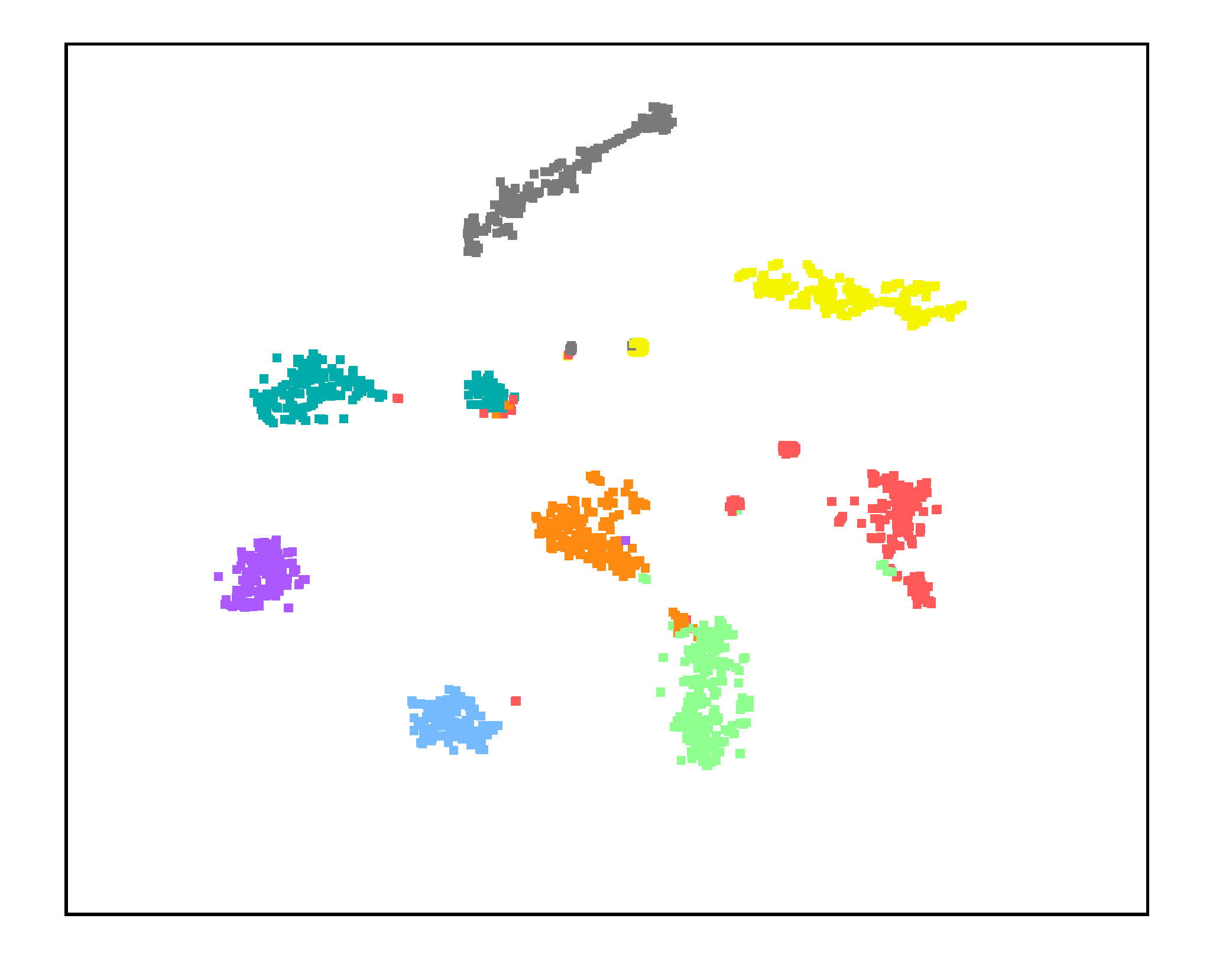}}
	\subfloat[CenGCN\_W]{\includegraphics[scale=0.22]{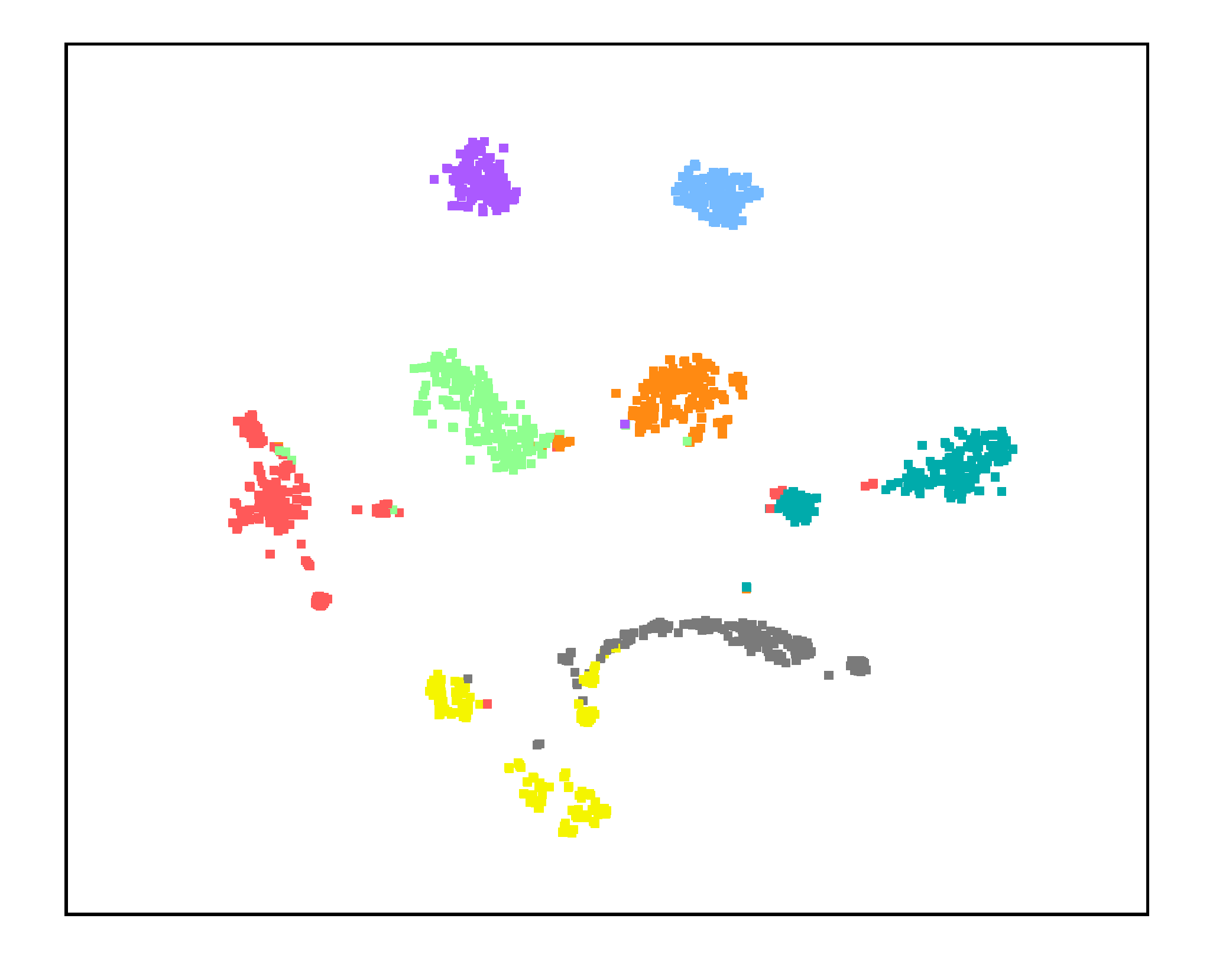}}
	\hfil
	\subfloat[CenGCN\_TD]{\includegraphics[scale=0.22]{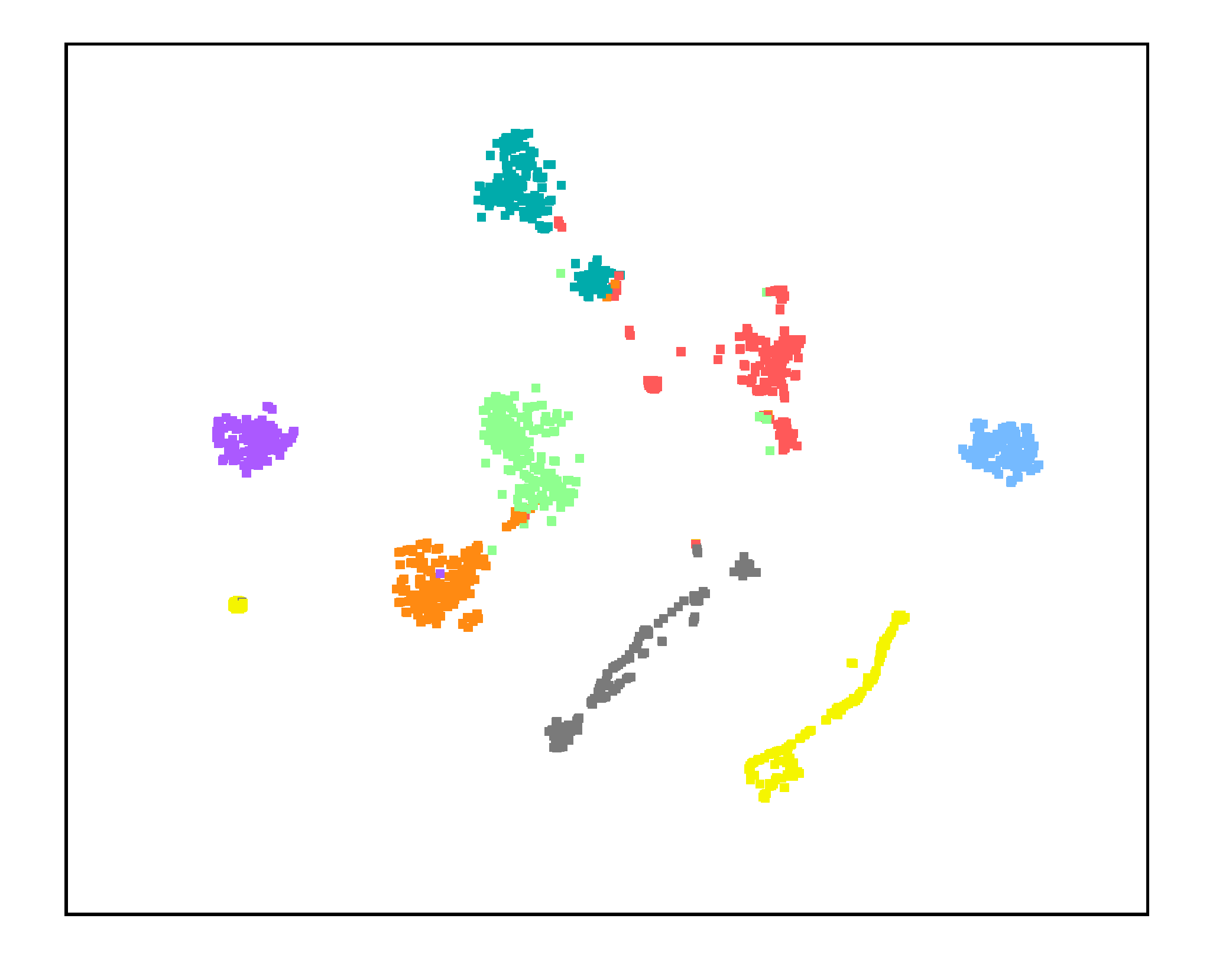}}
	\subfloat[CenGCN\_AD]{\includegraphics[scale=0.22]{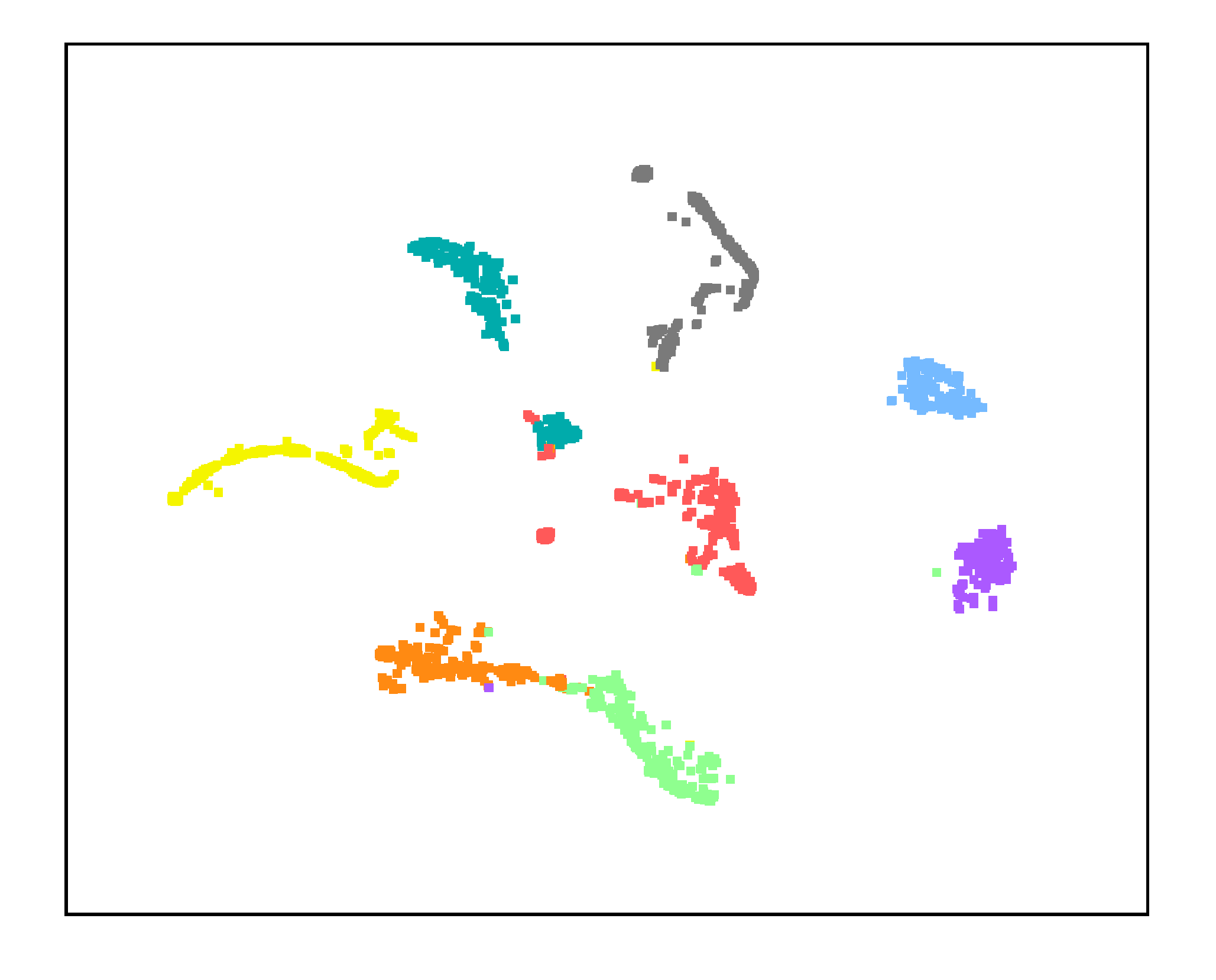}}
	\subfloat[CenGCN\_WD]{\includegraphics[scale=0.22]{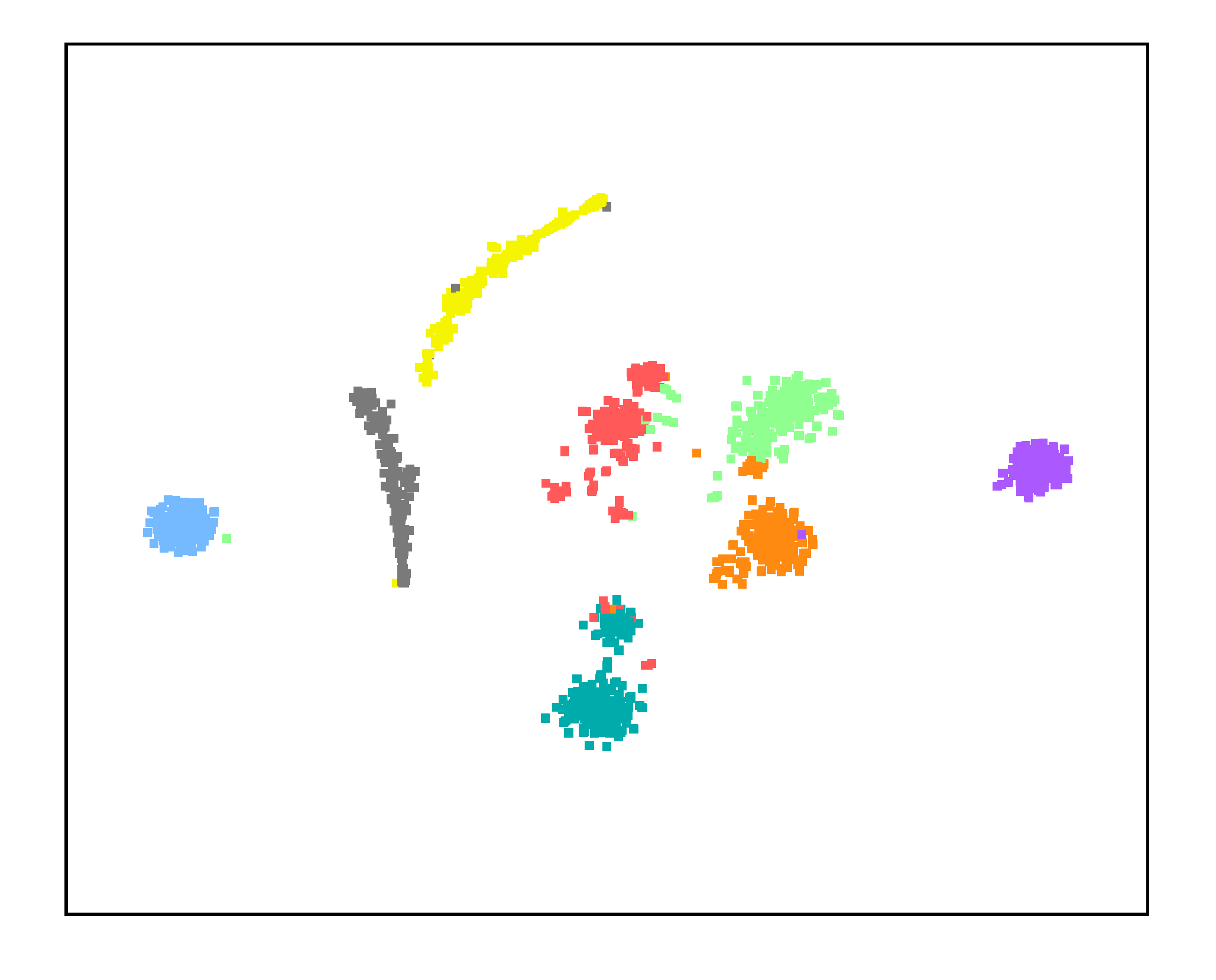}}
	\subfloat[CenGCN\_ID]{\includegraphics[scale=0.22]{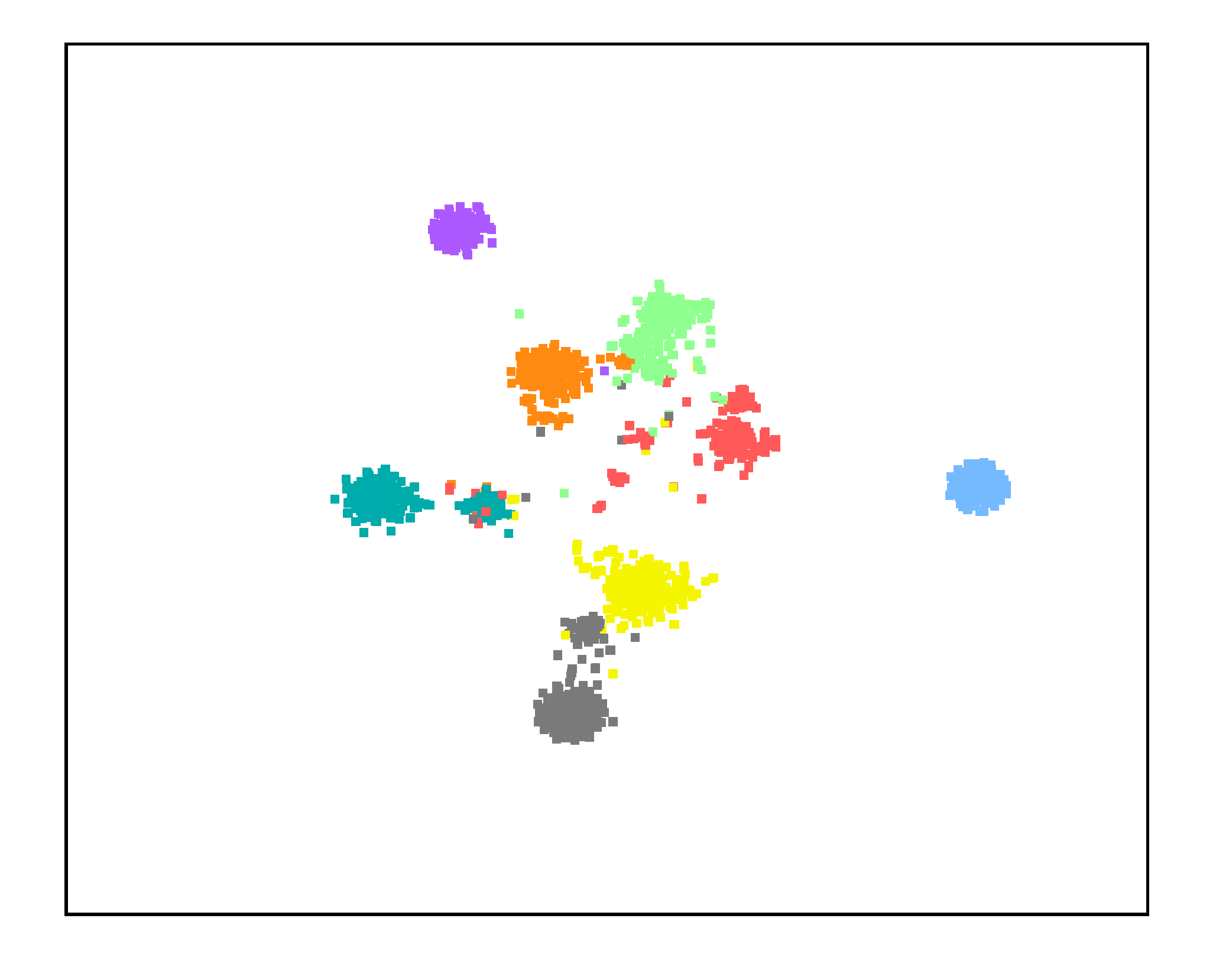}}
	\hfil
	\subfloat[CenGCN\_TE]{\includegraphics[scale=0.22]{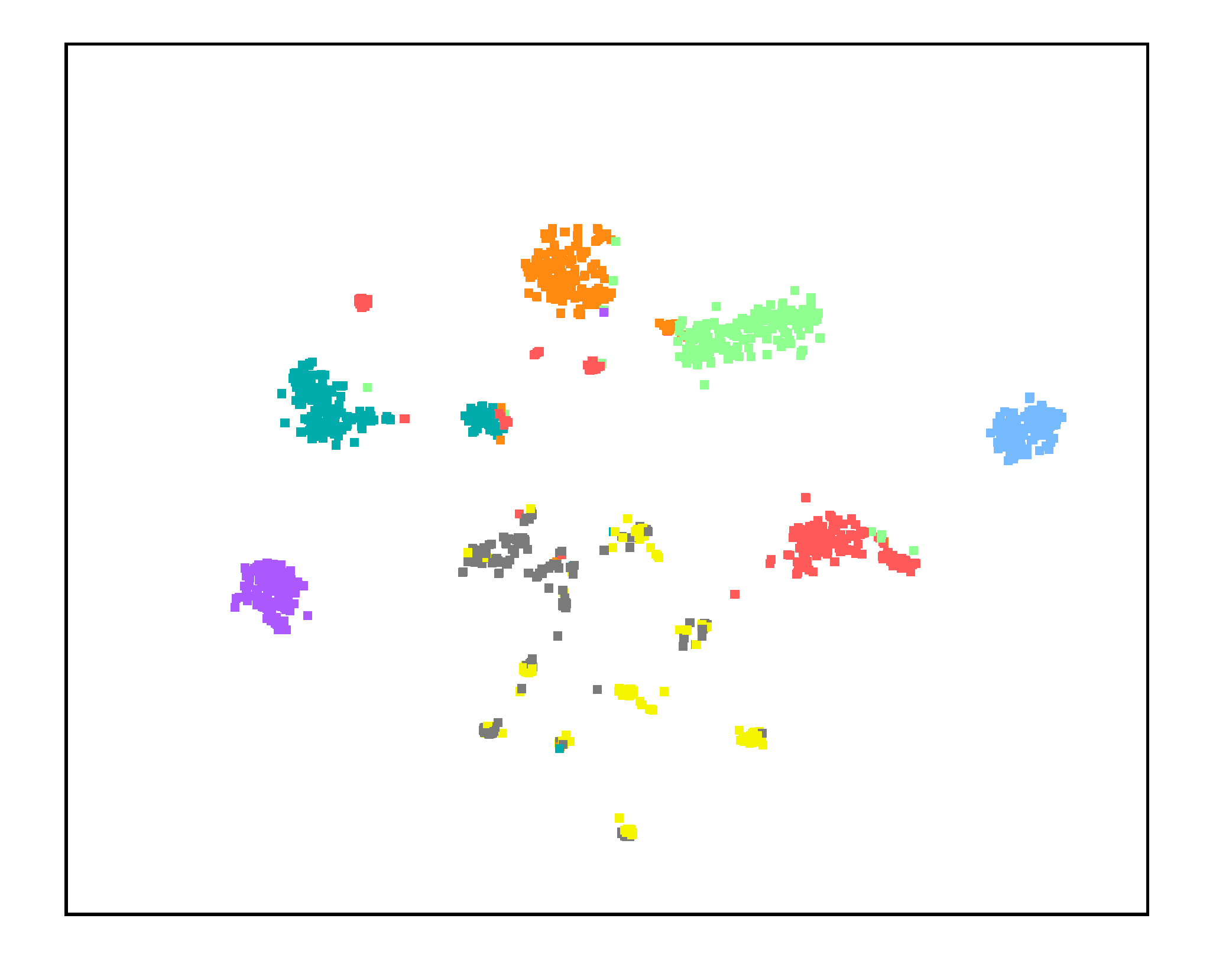}}
	\subfloat[CenGCN\_AE]{\includegraphics[scale=0.22]{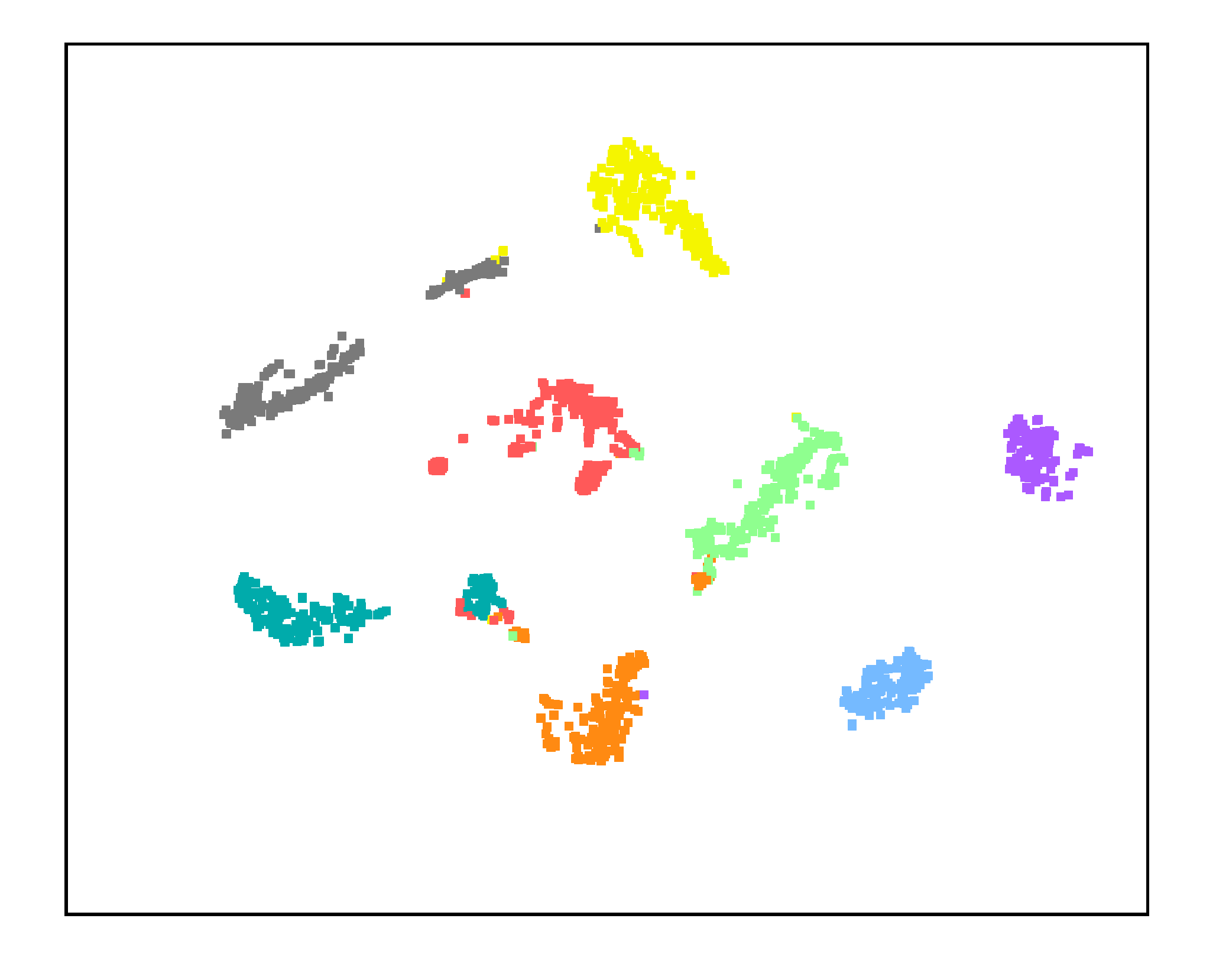}}
	\subfloat[CenGCN\_WE]{\includegraphics[scale=0.22]{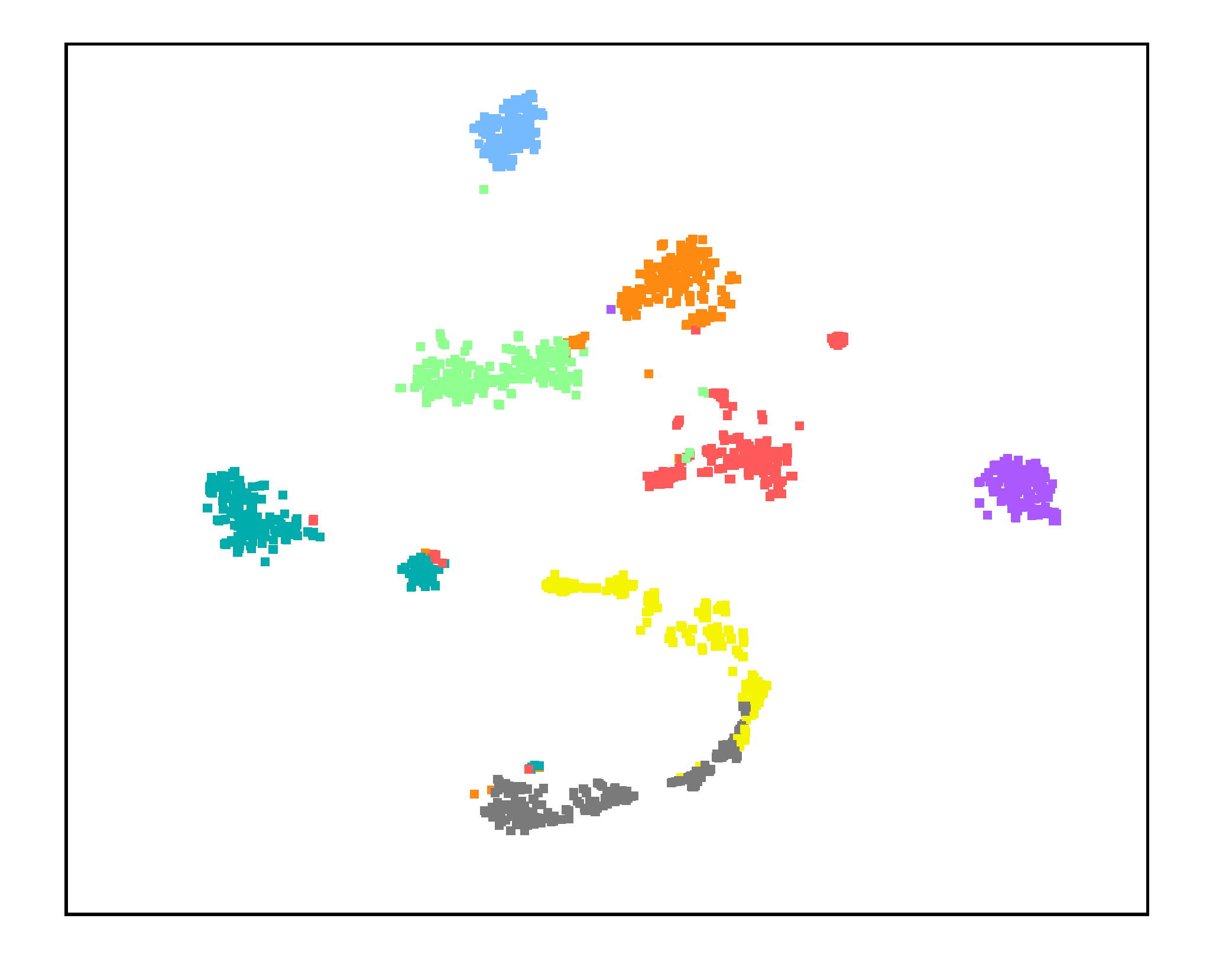}}
	\subfloat[CenGCN\_IE]{\includegraphics[scale=0.22]{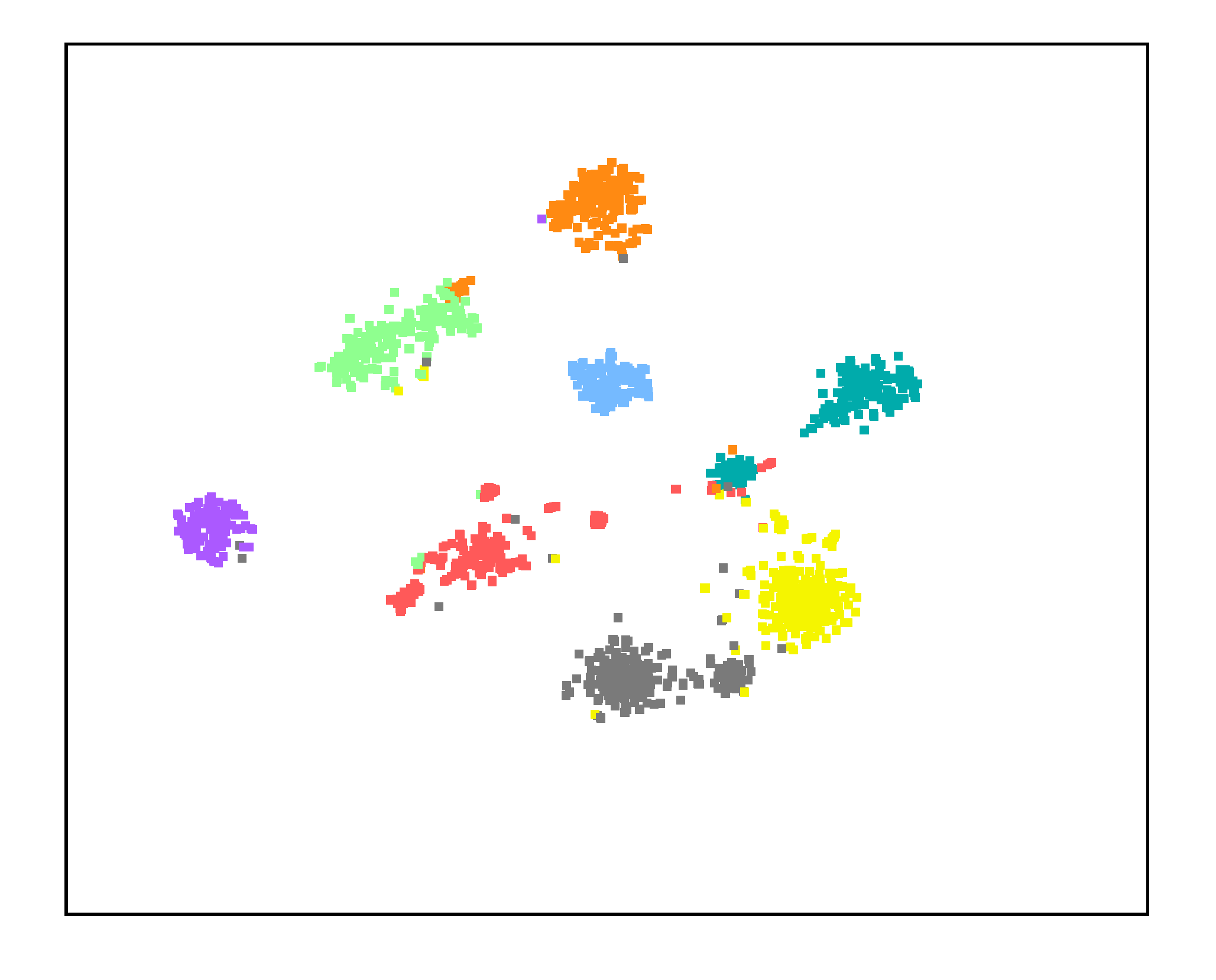}}
	\caption{Network Visualization using t-sne. Each color represents one class.}\label{figure:network_vis}
\end{figure*}

\subsection{Vertex clustering}
Vertex clustering is a typical unsupervised learning task and is used to find which vertices form a group. In this experiment, we compare the performance of CenGCN and baselines through this task.
We show the NMI in Table \ref{table:clustering}, where the best performance is reported by boldface. Table \ref{table:clustering} shows that on the five networks, CenGCN\_D achieves the best performance and CenGCN\_W performs better than all baselines, further suggesting the importance to utilise vertex centrality for GCNs.  Other noticeable observations from this experiment are summarized as follows:
\begin{itemize}
	\item CenGCN\_D always outperforms CenGCN\_E on the five networks. On Gplus, the NMI gap between them is 22.2\%. These results may be explained by the fact that the scale-free property is defined based on vertex degrees. When no class information is provided, network structures play a major role in training CenGCN.
	\item From the table, we can see that the two variants of CenGCN, CenGCN\_D and CenGCN\_E, outperform all baselines. CenGCN\_D achieves significant improvement on Facebook and Gplus. On Facebook, CenGCN\_D achieves improvement of at least 20.2\%; On Gplus, it achieves improvement of at least 27.7\%. Besides, CenGCN\_E achieves improvement of at least 10.8\% on Twitter.
	\item It still can be seen that CenGCN\_D performs best among variants with degree centrality and  CenGCN\_E performs best among variants with eigenvector centrality. The result further indicates the necessity of each part of CenGCN. It is surprising to find that only using the transformed graph or the hub attention is sufficient to achieve significant performance when we use degree centrality.
\end{itemize}

\subsection{Network Visualization}
\begin{figure}[htbp]
	\centering
	\includegraphics[scale=0.3]{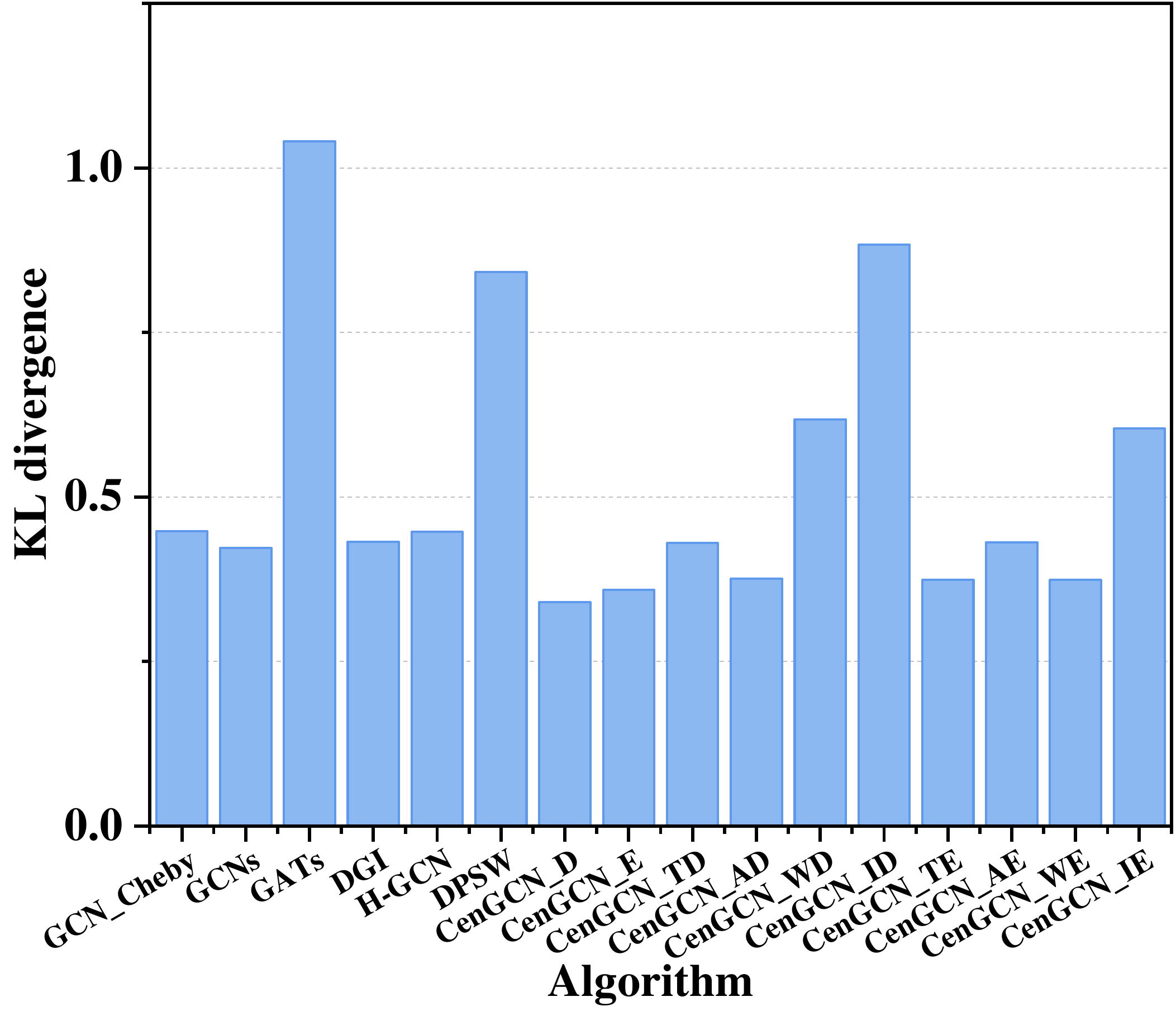}
	\caption{KL divergence on Twitter}
	\label{figure_kl}
\end{figure}

Network Visualization helps us explore the network structure in a low-dimensional space. In this experiment, we visualize the Twitter network using the learned vector representations. Fig. \ref{figure:network_vis} shows the visualized network on a 2-dimensional space, where each color represents one class. We summarize observable findings as follows:
\begin{itemize}
	\item GCN\_Cheby strongly confuses Red, Orange, and Light Green, and cannot develop boundaries to separate them. GCNs and GATs tightly connect Red, Orange, and Light Green, as well as Grey and Yellow. DGI poorly separates two subgroups of Grey. H-GCN poorly separates two subgroups of Sea Green. DPSW is insufficient to separate points of different colors. Also the points of the same color are less close together.
	\item CenGCN\_D shows the significant capacity of visualizing the Twitter network, sufficient to separate points of different colors and tightly cluster the points of the same color. CenGCN\_E also shows the significant capacity of visualizing this network, but it slightly confuses Yellow and Grey. Thus, these results suggest the usefulness of vertex centrality for network visualization.
	\item From the figure, we can see that some complementary variants of CenGCN also show significant performance in this experiment. Examples include CenGCN\_AD and CenGCN\_AE. But CenGCN\_AD slightly confuses Orange and Light Green. It can be seen that CenGCN\_TE fails to separate Green and Grey.
\end{itemize}
For quantitative comparison, we report KL divergences of algorithms in Fig. \ref{figure_kl}. KL divergences capture
the errors between the input pairwise similarities and their
projections in the 2-dimensional mapping. A lower KL divergence score indicates a better performance. We can see that CenGCN\_D and CenGCN\_E achieve the two smallest scores. Thus, they demonstrate better visualization performance than baselines.

\subsection{Parameter Sensitivity}
For CenGCN, three crucial hyper-parameters are $p$, $q$, and $r$. In this section, we investigate how they affect the performance of CenGCN on vertex classification. In addition to this, we also investigate the influence of the number of layers. For simplicity, we run experiments on CenGCN\_D and CenGCN\_E, omitting complementary variants.
\subsubsection{\textbf{The ratio of hub vertices}}

\begin{figure}[htbp]
	\centering
	\subfloat[CenGCN\_D]{
		\includegraphics[scale=0.2]{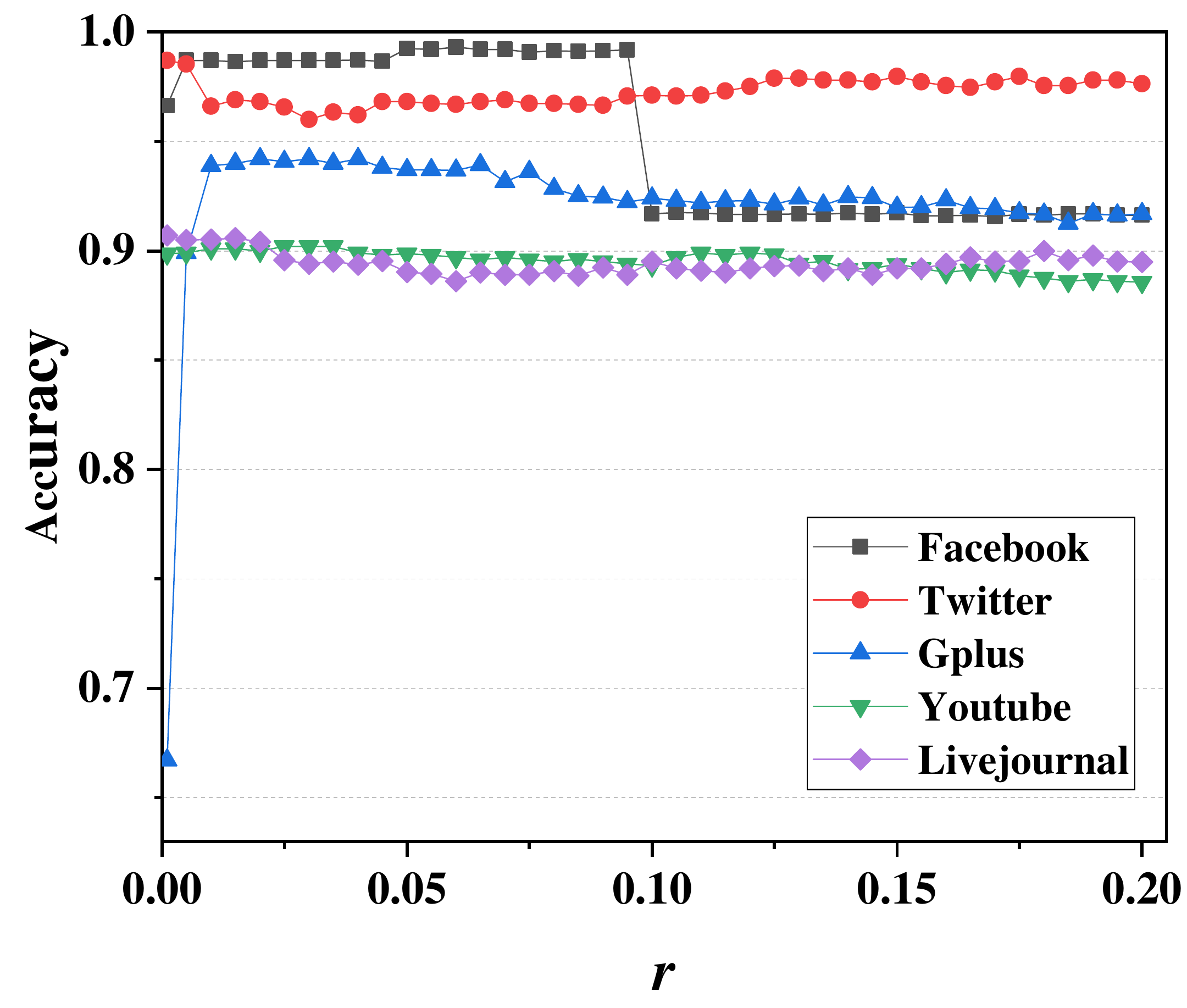}
		\label{r_a}
	}
	\subfloat[CenGCN\_E]{
		\includegraphics[scale=0.2]{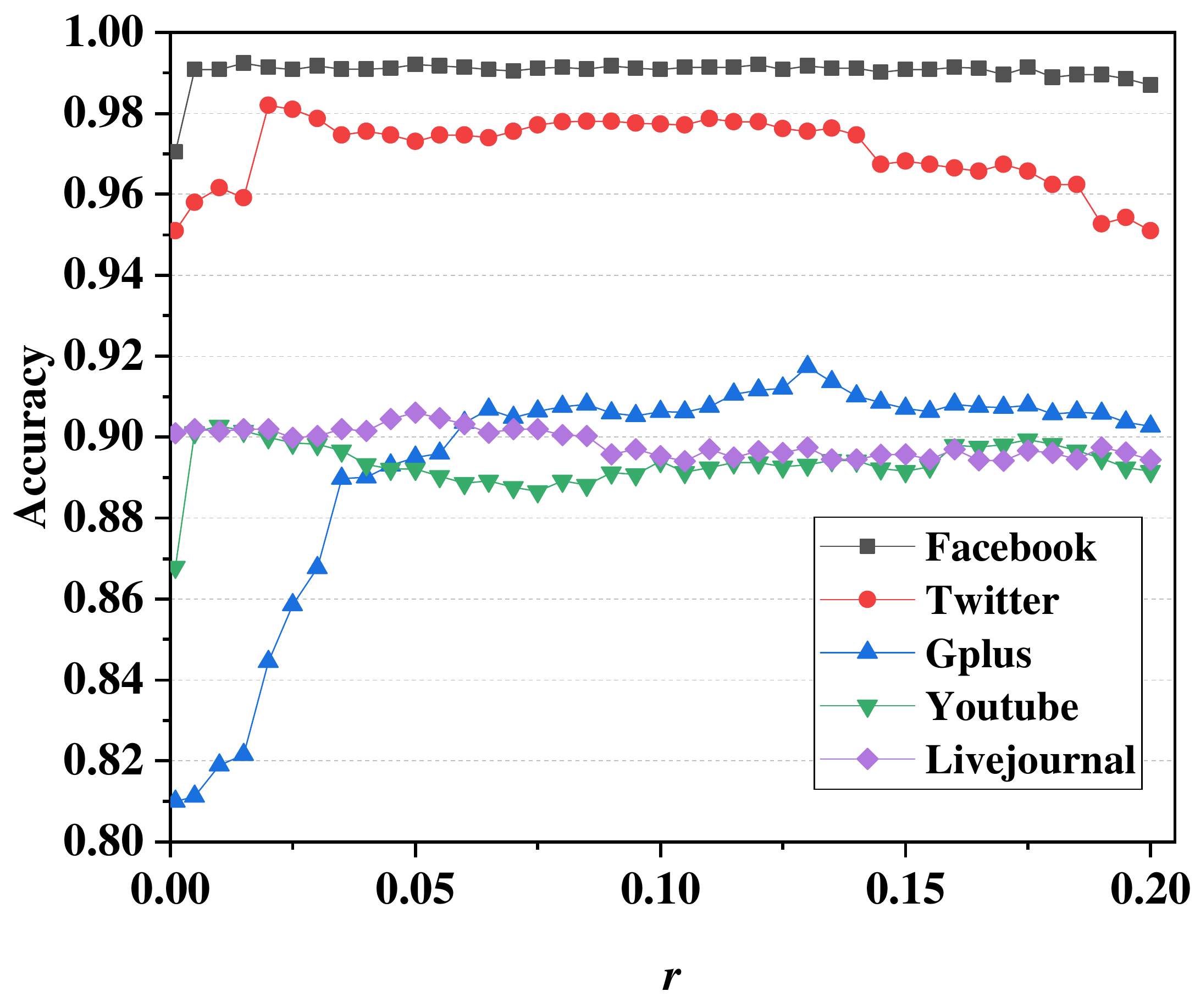}
		\label{r_b}
	}
	\caption{Sensitivities w.r.t. $r$}
	\label{figure:sen_r}
\end{figure}
\begin{figure}[htbp]
	\centering
	\subfloat[$p$]{
		\includegraphics[scale=0.2]{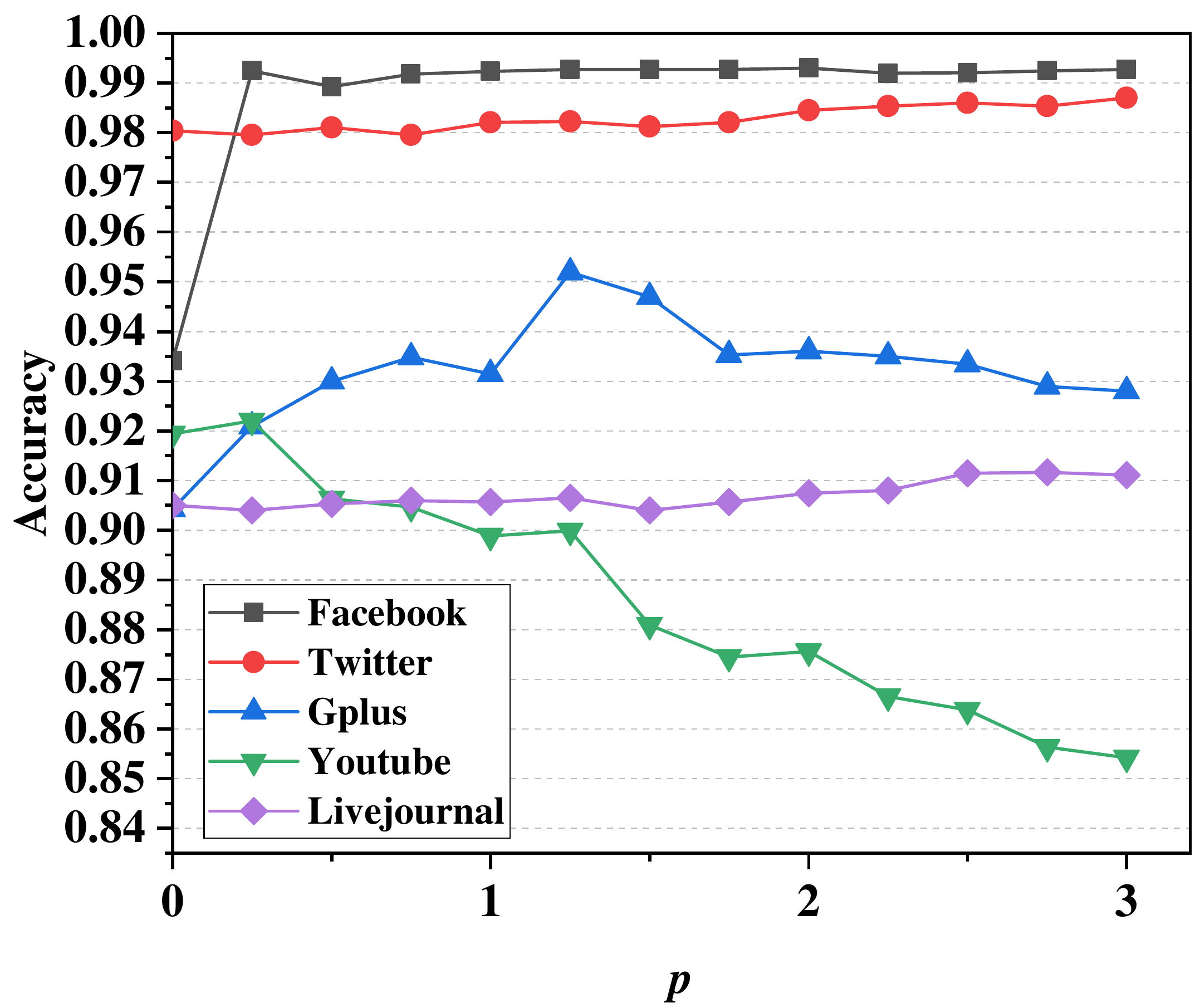}
		\label{p_d}
	}
	\subfloat[$q$]{
		\includegraphics[scale=0.2]{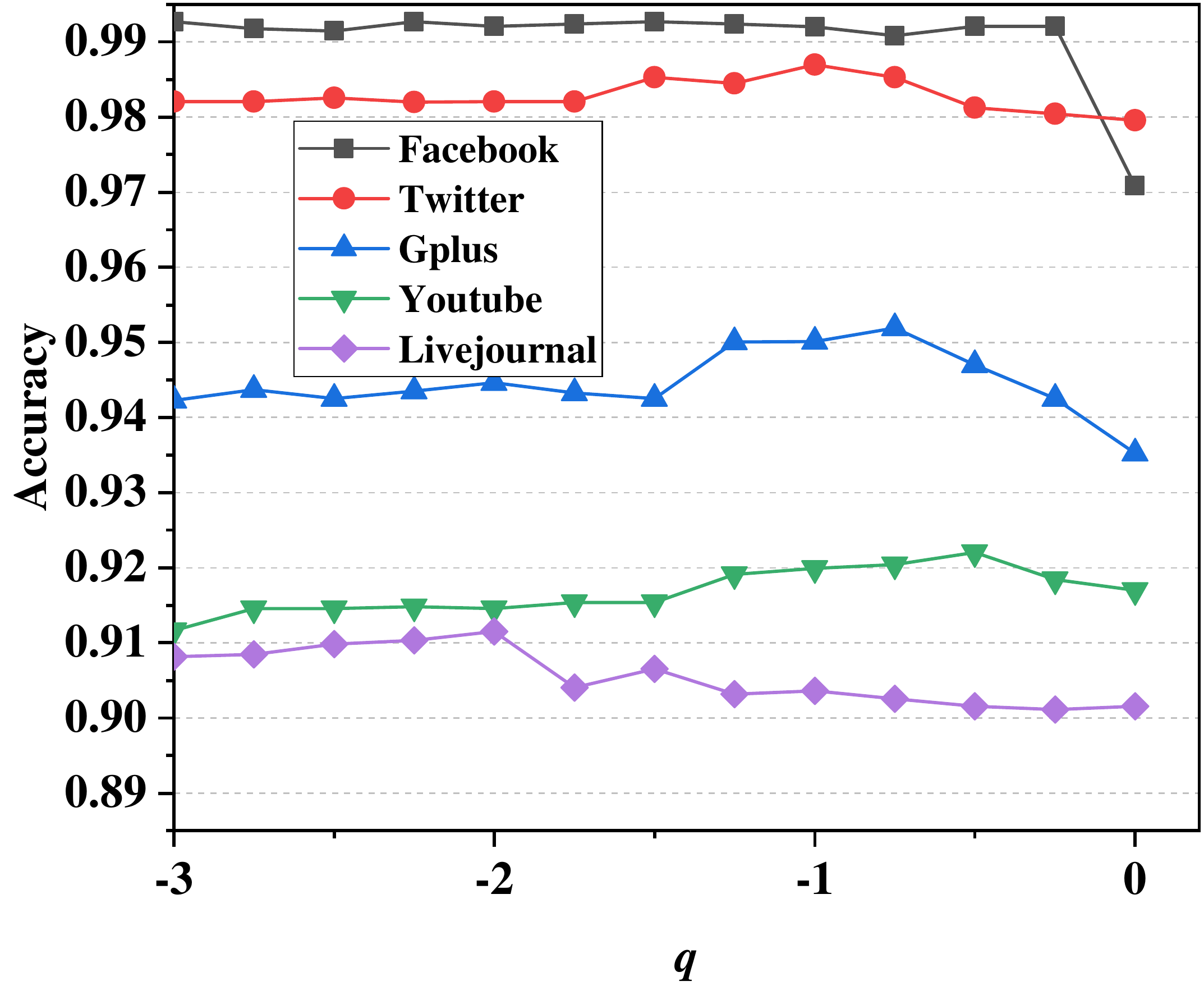}
		\label{q_d}
	}
	\caption{Sensitivities of CenGCN\_D w.r.t. $p$ and $q$}
	\label{figure:pq_d}
\end{figure}

\begin{figure}[htbp]
	\centering
	\subfloat[$p$]{
		\includegraphics[scale=0.2]{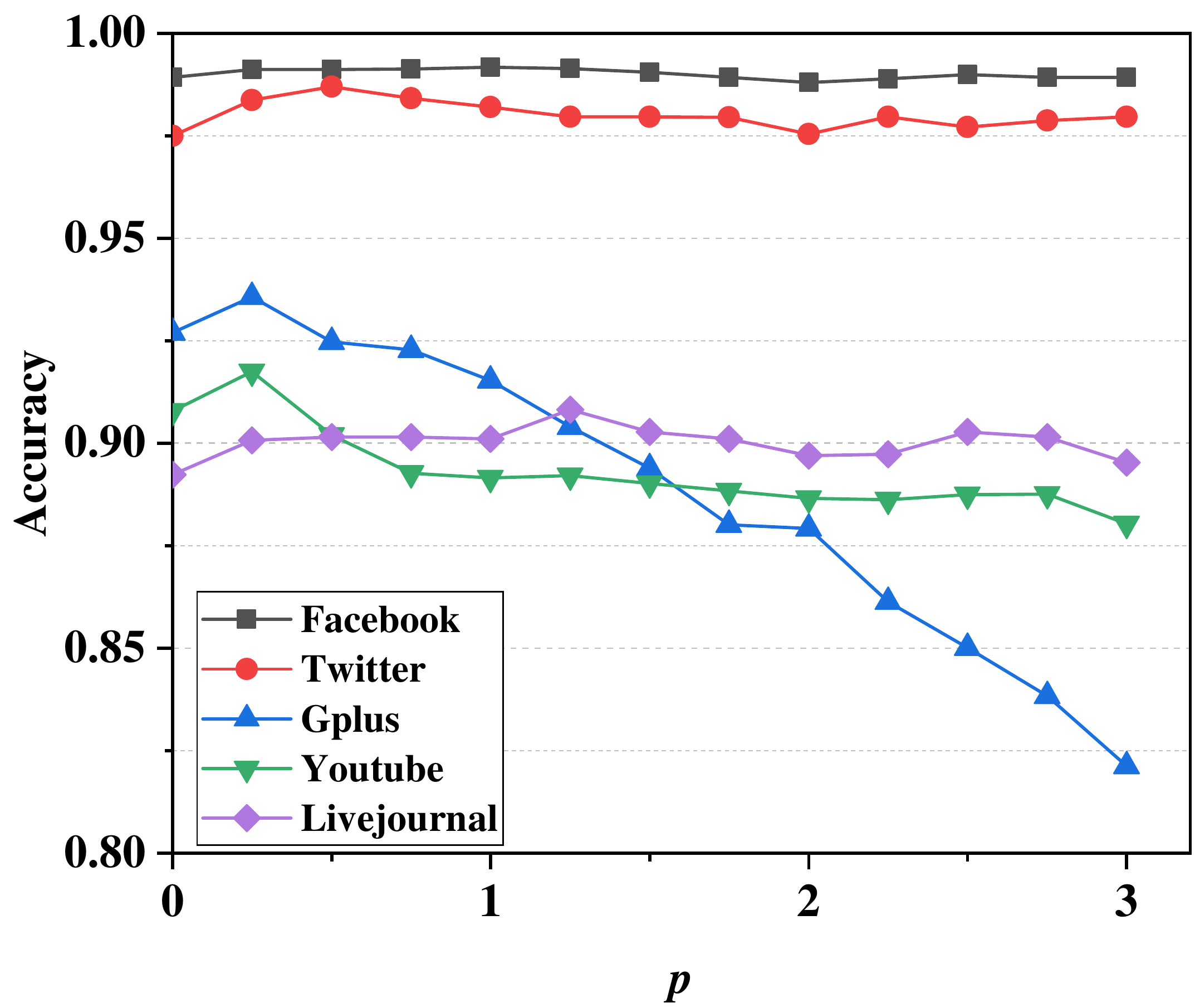}
		\label{p_e}
	}
	\subfloat[$q$]{
		\includegraphics[scale=0.2]{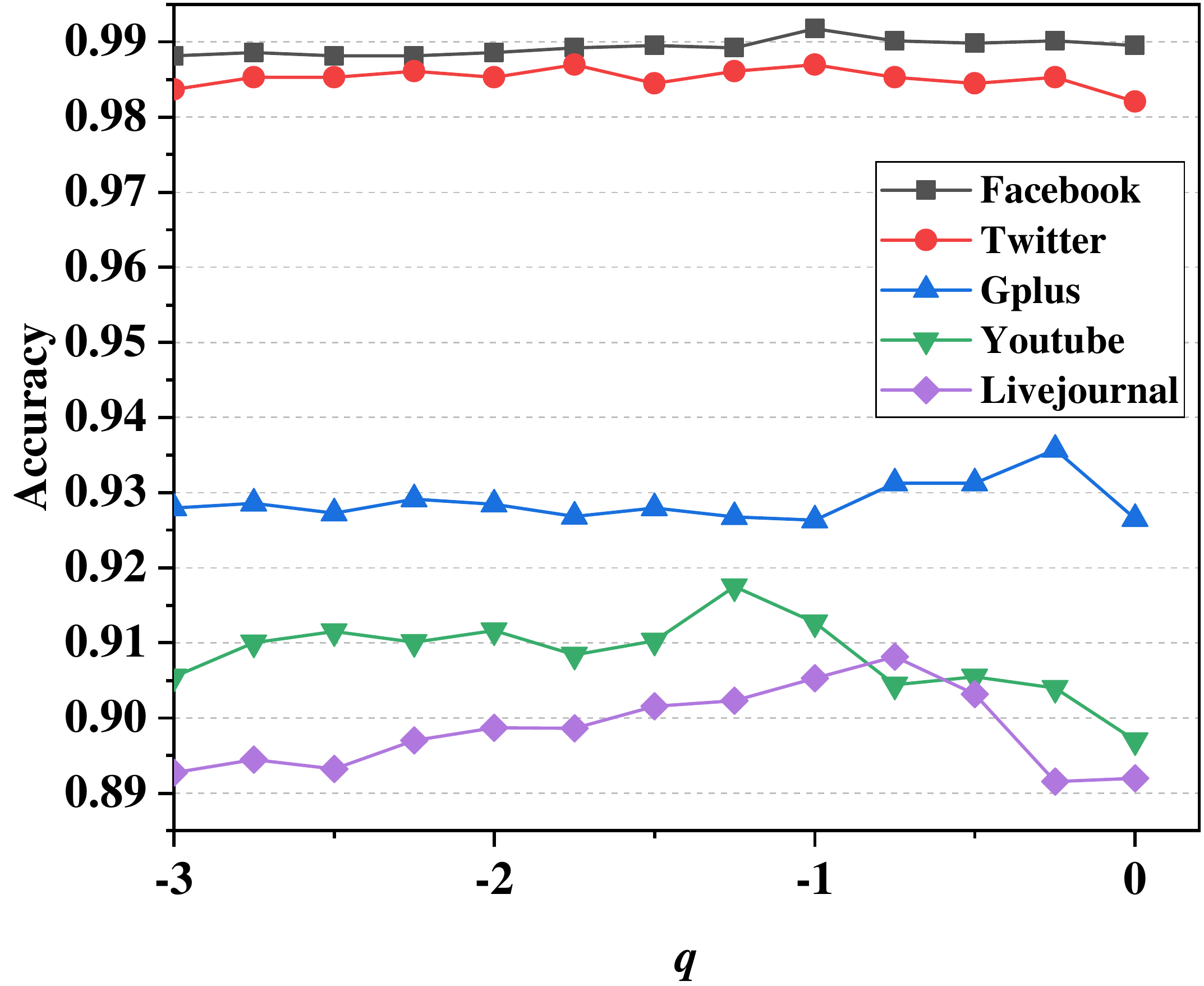}
		\label{q_e}
	}
	\caption{Sensitivities of CenGCN\_E w.r.t. $p$ and $q$}
	\label{figure:pq_e}
\end{figure}

\begin{figure*}[t]
	\centering
	\subfloat[Facebook]{\includegraphics[scale=0.17]{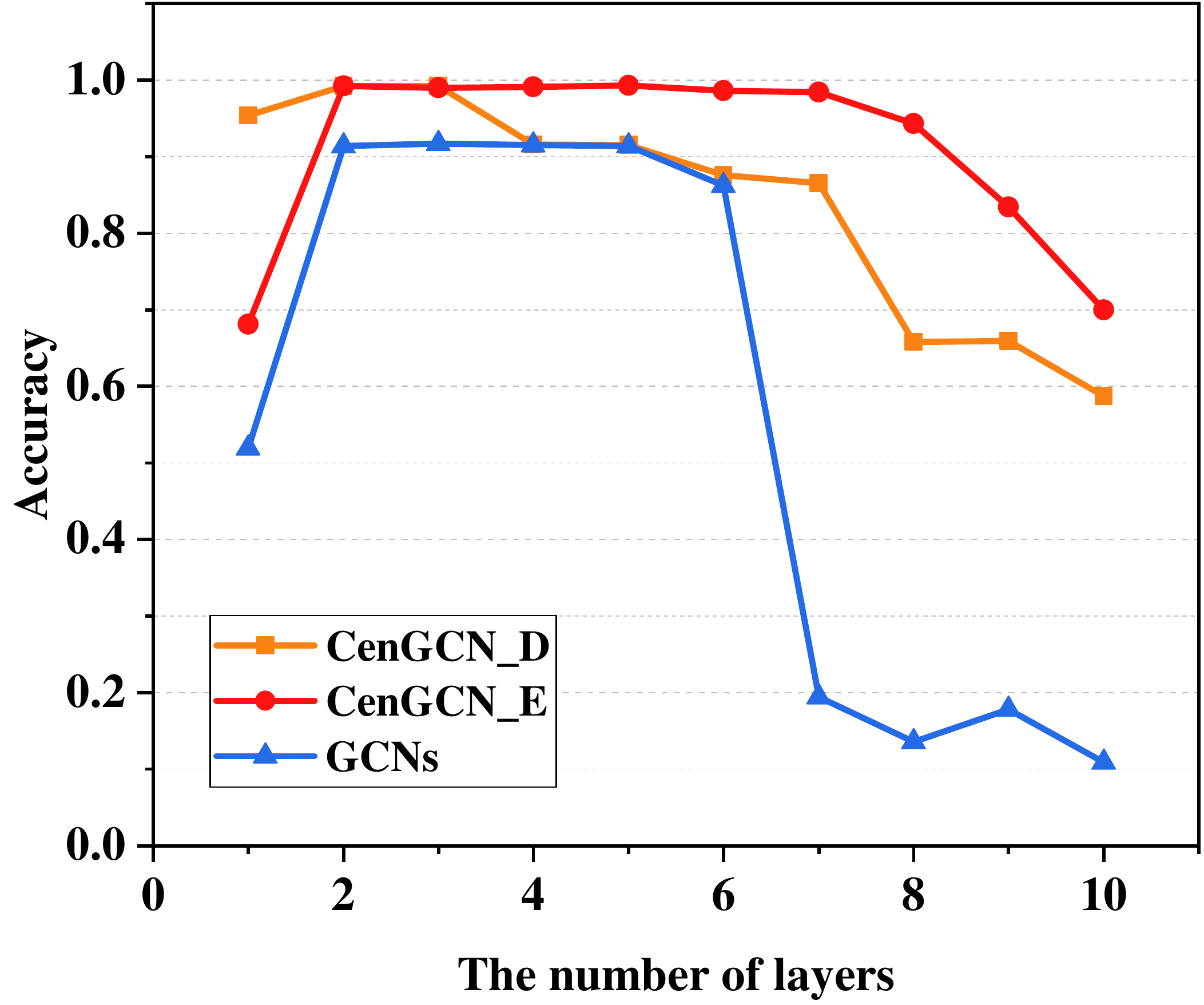}}
	\subfloat[Twitter]{\includegraphics[scale=0.17]{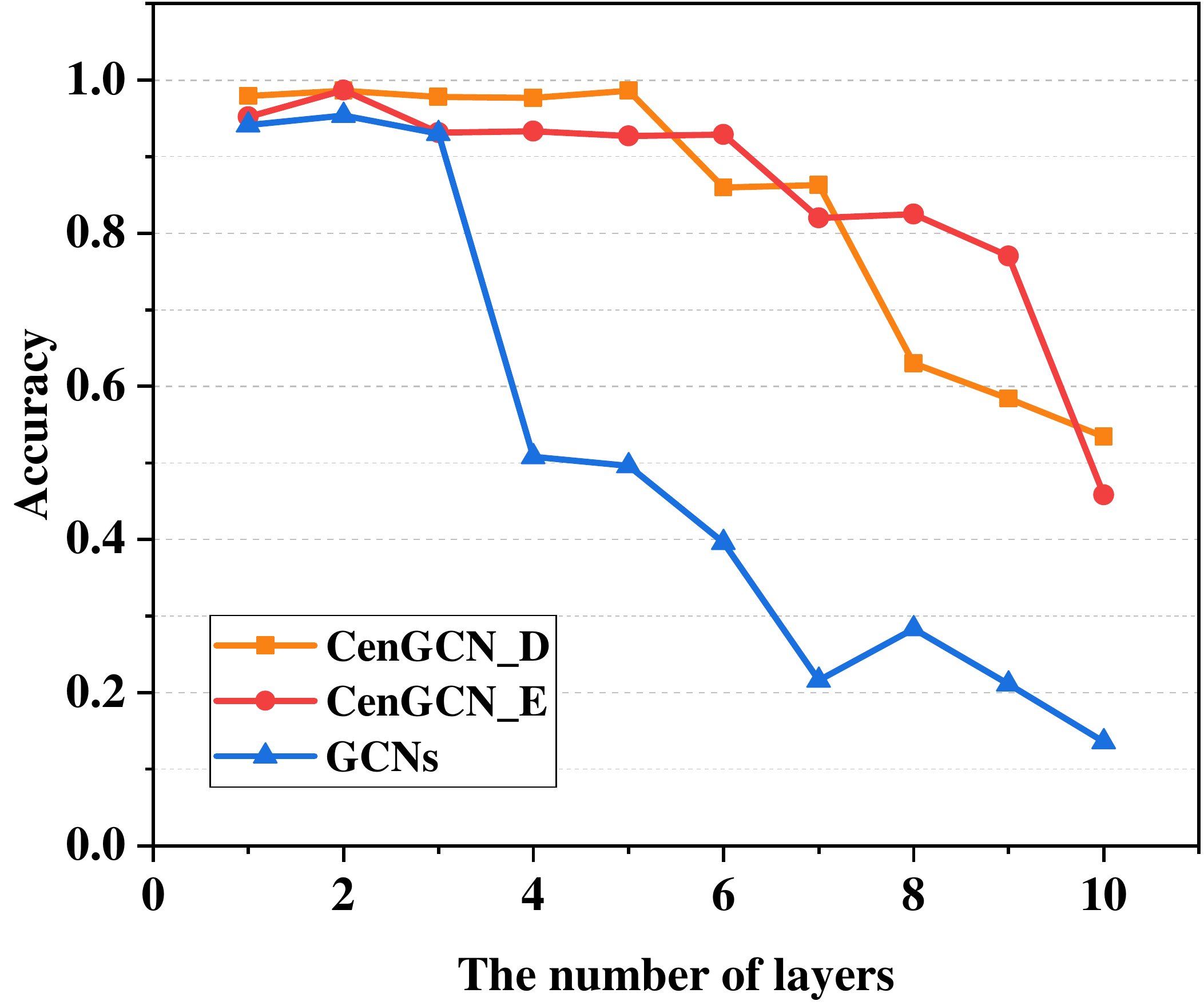}}
	\subfloat[Gplus]{\includegraphics[scale=0.17]{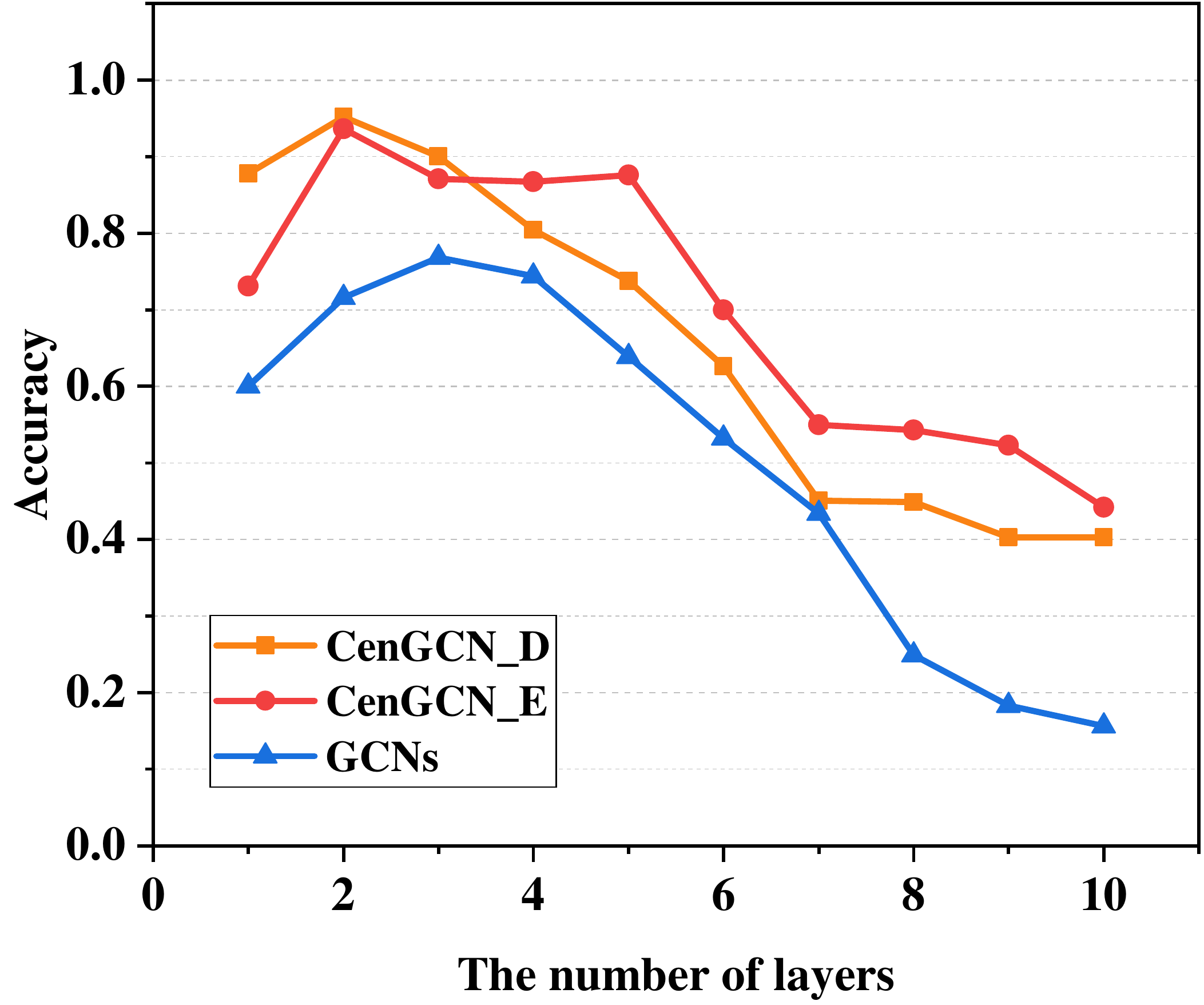}}
	\subfloat[Youtube]{\includegraphics[scale=0.17]{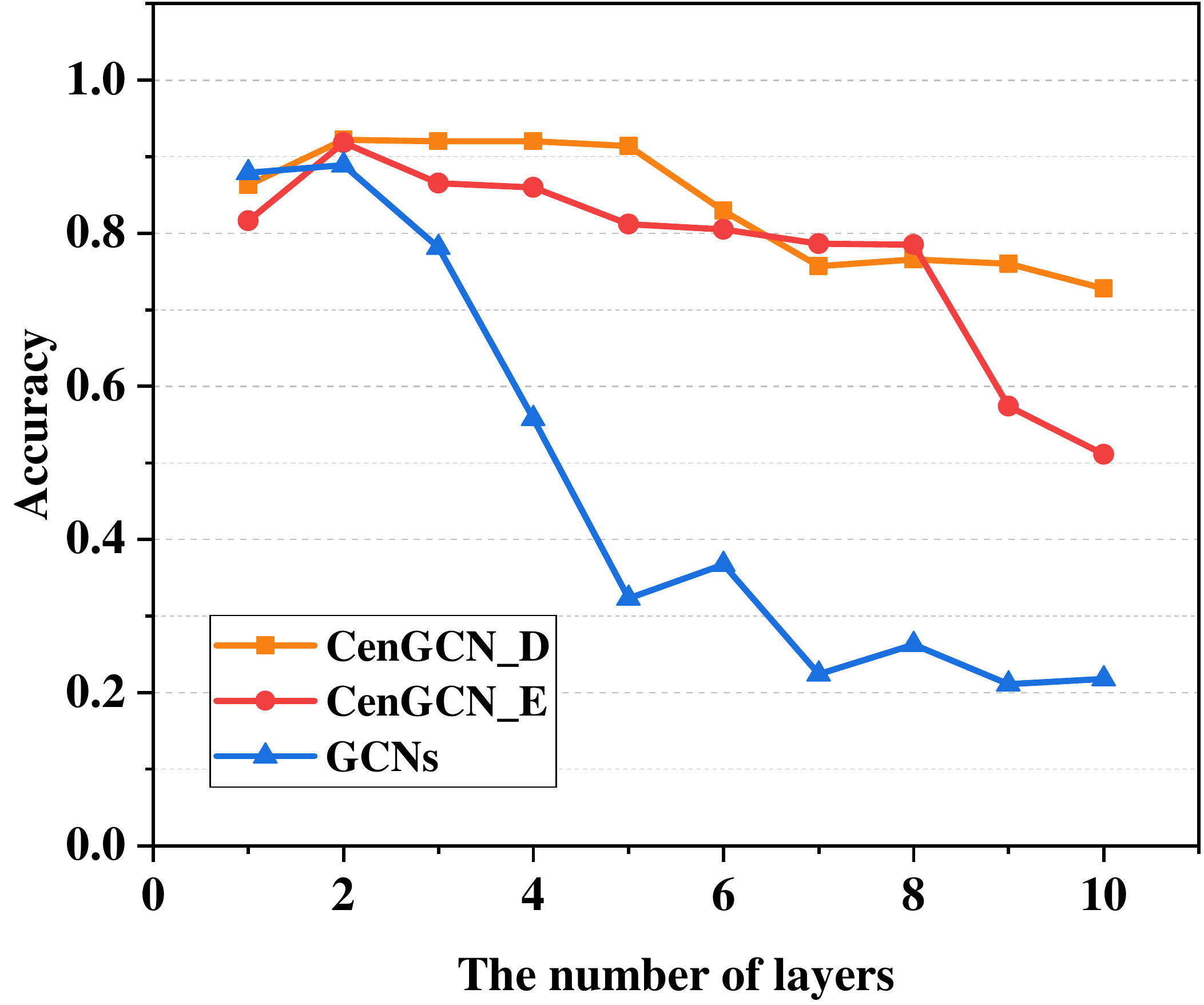}}
	\subfloat[Livejournal]{\includegraphics[scale=0.17]{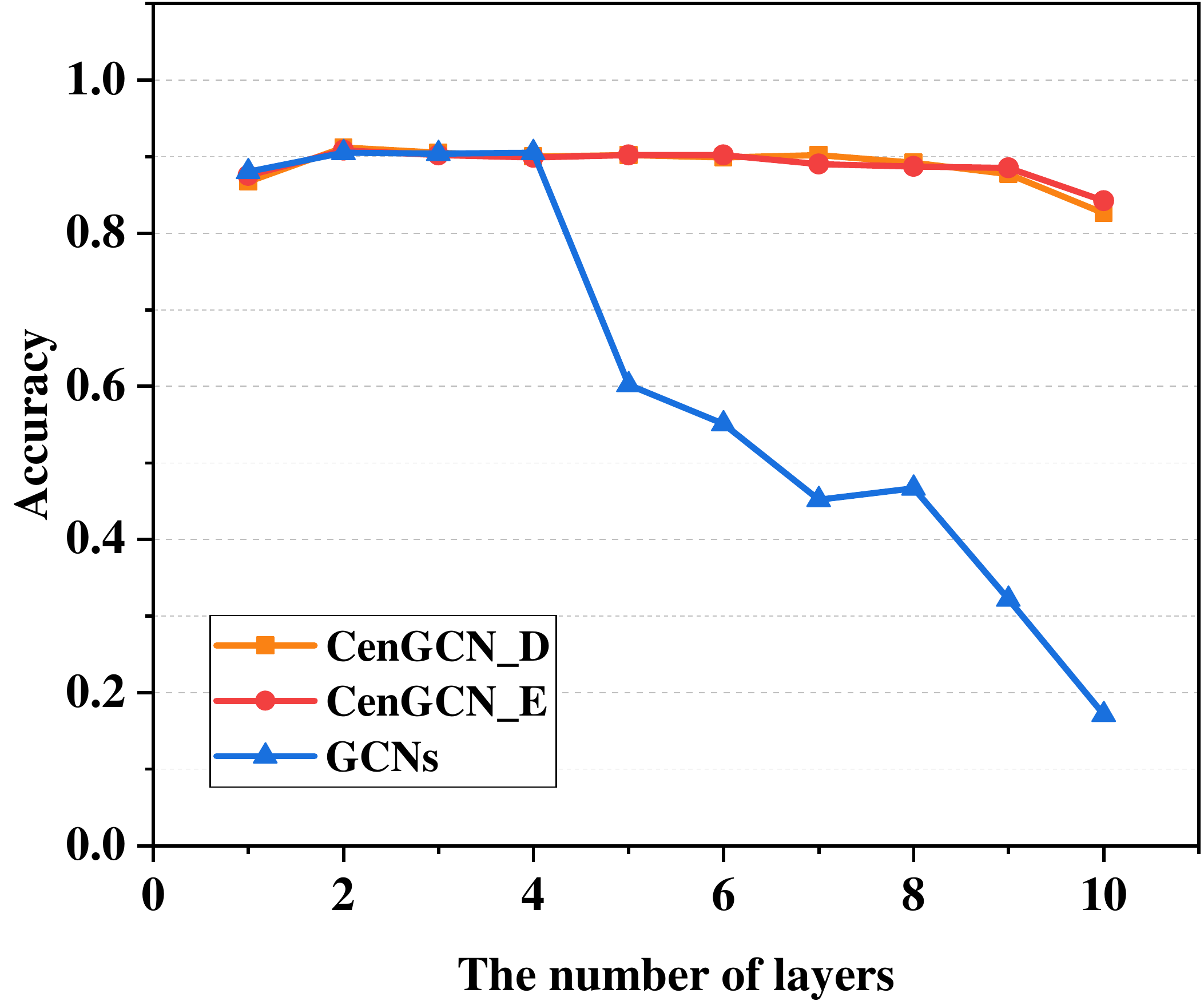}}
	\caption{Sentivities w.r.t. the number of layers} \label{figure:layer_num}
\end{figure*}

We select $r$ in the range from 0.001 to 0.2. A bigger $r$ implies more hub vertices to be considered. Here, we set $p$ and $q$ to 1 and -1, respectively. The Fig. \ref{figure:sen_r} shows sensitivities of CenGCN\_D and CenGCN\_E w.r.t. $r$. From the figure, we can see different variation tendencies in different networks, as $r$ goes from 0.001 to 0.2. We summarize findings in the five datasets as follows:
\begin{itemize}
	\item On Facebook, accuracies of CenGCN\_D and CenGCN\_E abruptly become higher when $r$ increases from 0.001 to larger values. CenGCN\_D maintains its peak from 0.05 to 0.095. After that, its accuracy drastically decreases to a low value, which is maintained as $r$ is from 0.1 to 0.2. CenGCN\_E maintains its peak when $r$ is between 0.005 and 0.165. Finally, its accuracy starts to slowly decrease.
	\item  On Twitter, the accuracy of CenGCN\_D immediately reaches its peak when $r$ passes from 0.001. It then starts to decrease and slowly increases when $r$ is over 0.05. The accuracy of CenGCN\_E reaches its peak when $r$ is 0.02. After that, it experiences a decrease, then an increase and finally a decrease.{\tiny }
	\item On Gplus, the accuracy of CenGCN\_D significantly increases when $r$ passes from 0.001 to 0.01. When $r$ continues to 0.03, the accuracy gently increases. Finally, it decreases to a stable value. The accuracy of CenGCN\_E maintains an increasing tendency until $r$ is up to 0.13. This tendency first is very strong and then becomes slow. After $r$ is over 0.13, the accuracy starts to decrease.
	\item On Youtube, the accuracy of CenGCN\_D has a stable change. It first gently increases until $r$ is around 0.03. Then it experiences a decrease, an increase and finally a decrease. The accuracy of CenGCN\_E has a drastic increase when $r$ goes from 0.001 to higher values. It reaches its peak when $r$ is 0.01. After that, it first decreases and then increases. Finally, it decreases again when $r$ is over 0.16.
	\item On Livejournal, the accuracy of CenGCN\_D immediately reaches its peak when $r$ is 0.001. As $r$ passes to 0.2, the accuracy first decreases to a stable value and then increases. The accuracy of CenGCN\_E reaches its peak as $r$ goes from 0.001 to 0.05. After that, it starts to decrease to a value of around 0.895 with small fluctuations.
\end{itemize}

\subsubsection{\textbf{The influence extent of vertex centrality}}
The two parameters $p$ $(>0)$ and $q$ $(<0)$ control the influence extent of vertex centrality. In this experiment, we investigate the sensitivities w.r.t. $p$ and $q$. Fig. \ref{figure:pq_d} and Fig. \ref{figure:pq_e} show the sensitivity results of CenGCN\_D and CenGCN\_E, respectively. From the two figures, we can observe the following findings:
\begin{itemize}
	\item On Facebook, the accuracy of CenGCN\_D maintains stability when $p$ and $q$ are non-zero values. For CenGCN\_E, its accuracy has a peak when $p$ is 1 and $q$ is -1. Outside of the peak, it also maintains a stable value.
	\item On Twitter, the accuracy of CenGCN\_D steadily and slowly increases as $p$ goes from 0 to 3. With $q$ increasing, the accuracy increases from a stable value to its peak, where $q$ is -1. For CenGCN, its accuracy is maximum when $p$=0.5. As $p$ becomes larger, the accuracy maintains stability. It is also a stable value when $q$ is over -3 and below 0.
	\item On Gplus, the accuracy of CenGCN\_D has a conspicuous peak when $p$=1.25 and $q$=-0.75. It is stable as $q$ goes from -3 to -1.5. Similar to CenGCN\_D, the accuracy of CenGCN\_E also has a conspicuous peak, where $p$=0.25 and $q$=-0.25. We can see that the accuracy decreases linearly when $p$ is over 0.25.
	\item On Youtube, CenGCN\_D needs a small $p$. As $p$ goes from 0.25 to 3, its accuracy steadily decreases. The accuracy slowly increases when $q$ increases to -0.5. CenGCN\_e also needs a small $p$, but it needs a smaller $q$ of -1.25.
	\item On Livejournal, the accuracy of CenGCN\_D maintains a stable increase as $p$ becomes larger, and maintains a stable decrease as $q$ goes from -2 to 0. The accuracy of CenGCN\_E has a peak when $p$=1.25. On either sides of this peak, it maintains stability. As $q$ goes from -3 to -0.75, the accuracy steadily increases. After that, it drastically decreases.
\end{itemize}

\subsubsection{\textbf{The number of layers}}
In this experiment, we investigate how the number of layers affects the performance of CenGCN\_D and CenGCN\_E. Besides, the performance of GCNs is added here for a comparison. The $p$ and $q$ are set to their optimal values. Fig. \ref{figure:layer_num} shows their performance in the five networks when the number increases from 1 to 10. We summarize noticeable findings as follows:
\begin{itemize}
	\item CenGCN\_D, CenGCN\_E, and GCNs achieve great performance when the number of layers is 2. One exception is that on Gplus, GCNs have the best performance when the number is 3. Therefore, it is reasonable for GCN-based models to design a two-layer neural network.
	\item When the number is greater than 2 and continues to increase, the accuracies of GCNs show decreasing tendencies. When the number increases to 10, the accuracy of GCNs is below or slightly over 0.2. The result shows that GCNs suffer from shallow models.
	\item As the number increases, the extent by which CenGCN\_D and CenGCN\_E outperform GCNs also increases. On LiveJournal, when the number is at 10, CenGCN\_D and CenGCN\_E achieve accuracies of more than 0.8, significantly outperforming GCNs. On the other four networks, we can also observe a large gap between the two variants of CenGCN and GCNs. These figures demonstrate that CenGCN deepens GCNs.
\end{itemize}

Existing GCN-based methods employ a two-layer neural network. Such a network merely mines the relationship between vertices whose distance is at most 2-hop, failing to exploit global network structure. How to deepen GCNs is still an open issue and worthwhile to study further. This experiment shows that CenGCN has excellent performance compared with GCNs when the number of layers increases to 10. Therefore, for deepening GCNs, a suggestion from this experiment is that we can utilise vertex centrality.

\section{Conclusion}
In this paper, we study how to address the inequality of information from vertices. We propose label propagation with labeled hub vertices to quantify the similarity between hub vertices and their neighbors. Based on this similarity and centrality indices, we transform the graph to capture the influence of hub vertices. When inputting the transformed graph into GCNs, we propose a hub attention mechanism to learn new weights linking to non-hub neighbors from the same hubs. In four experiments, the two variants, CenGCN\_D and CenGCN\_E, demonstrate their significant improvement over baselines and excellent performance when the number of layers increases to 10.

GCNs are rapidly developing and proving effective tools in network modeling and analysis. Although there are many studies about GCNs, a great number of issues remain to be addressed. Two serious issues are that GCNs suffer from local limits and shallow models. This study demonstrates a way to explore vertex imbalance and unequal information by vertex centrality, a macroscopic network characteristic, to enhance and enrich GCNs. In the future, we will consider more network characteristics, such as subgraphs.

% \begin{appendix}
% 	This supplementary material provides more detailed experimental results, including standard deviations calculated for multiple experiments.
% 	section
% \end{appendix}

% or


%\appendices
%% you can choose not to have a title for an appendix
%% if you want by leaving the argument blank
%\section{}

%\ifCLASSOPTIONcompsoc
%  % The Computer Society usually uses the plural form
%\else
%  % regular IEEE prefers the singular form
%\fi

\ifCLASSOPTIONcaptionsoff
  \newpage
\fi



%\bibliographystyle{IEEEtran}
%\begin{thebibliography}{1}
%\end{thebibliography}
\bibliographystyle{IEEEtran}
\bibliography{IEEEabrv,reference}

\appendix[Supplementary Experimental Results]
% \section*{add}
This supplementary material provides more detailed experimental results, including standard deviations calculated for multiple experiments. 
\par On the vertex classification and link prediction tasks, we perform several experiments to calculate the standard deviation of the method as a way of verifying the robustness of the model. Where DPSW is not affected by random seeds and therefore has a standard deviation of 0. The specific results are shown in Table \ref{table:classification2}, \ref{table:link2}.
\begin{table*}[htbp]
	
	\setlength\tabcolsep{9pt}
	\centering
	\caption{ The accuracy of vertex classification.} \label{table:classification2}
	\begin{tabular}{c|c|c|c|c|c}
		\toprule
		Algorithm &  Facebook & Twitter & Gplus & Youtube & LiveJournal\\
		\midrule
		GCN\_Cheby & 0.915 $\pm$ 0.0018 & 0.972 $\pm$ 0.0029 & 0.787 $\pm$ 0.0062 & 0.812 $\pm$ 0.0077 & 0.810 $\pm$ 0.0071 \\
		GCNs & 0.914 $\pm$ 0.0018 & 0.954 $\pm$ 0.0110 & 0.716 $\pm$ 0.0101 & 0.889 $\pm$ 0.0146 & 0.901 $\pm$ 0.0042 \\
		GATs & 0.970 $\pm$ 0.0016 & 0.967 $\pm$ 0.0036 & 0.732 $\pm$ 0.0227 & 0.827 $\pm$ 0.0028 & 0.892 $\pm$ 0.0080 \\
		DGI & 0.936 $\pm$ 0.0013 & 0.954 $\pm$ 0.0023 & 0.771 $\pm$ 0.0087 & 0.227 $\pm$ 0.0015 & 0.592 $\pm$ 0.0136 \\
		H-GCN & 0.982 $\pm$ 0.0015 & 0.943 $\pm$ 0.0045 & 0.914 $\pm$ 0.0032 & 0.915 $\pm$ 0.0045 & 0.888 $\pm$ 0.0011 \\
		DPSW & 0.892 $\pm$ 0. & 0.789 $\pm$ 0. & 0.922 $\pm$ 0. & 0.892 $\pm$ 0. & 0.872 $\pm$ 0. \\
		\midrule
		CenGCN\_D & \textbf{0.992 $\pm$ 0.0014} & \textbf{0.987 $\pm$ 0.0010} & \textbf{0.949 $\pm$ 0.0017} & \textbf{0.920 $\pm$ 0.0013} & 0.912 $\pm$ 0.0015 \\
		CenGCN\_TD & 0.970 $\pm$ 0.0018 & 0.982 $\pm$ 0.0120 & 0.943 $\pm$ 0.0033 & 0.914 $\pm$ 0.0016 & 0.897 $\pm$ 0.0018 \\
		CenGCN\_AD & 0.970 $\pm$ 0.0012 & 0.969 $\pm$ 0.0121 & 0.933 $\pm$ 0.0026 & 0.915 $\pm$ 0.0025 & 0.910 $\pm$ 0.0045 \\
		CenGCN\_WD & 0.832 $\pm$ 0.0012 & 0.965 $\pm$ 0.0015 & 0.941 $\pm$ 0.0019 & 0.904 $\pm$ 0.0018 & 0.903 $\pm$ 0.0018 \\
		CenGCN\_ID & 0.888 $\pm$ 0.0024 & 0.967 $\pm$ 0.0018 & 0.861 $\pm$ 0.0044 & 0.893 $\pm$ 0.0027 & 0.893 $\pm$ 0.0020 \\
		\midrule
		CenGCN\_E & \textbf{0.992 $\pm$ 0.0015} & \textbf{0.987 $\pm$ 0.0012} & 0.936 $\pm$ 0.0018 & 0.919 $\pm$ 0.0015 & \textbf{0.903 $\pm$ 0.0030} \\
		CenGCN\_TE & 0.912 $\pm$ 0.0019 & 0.905 $\pm$ 0.0010 & 0.717 $\pm$ 0.0014 & 0.873 $\pm$ 0.0028 & 0.894 $\pm$ 0.0017 \\
		CenGCN\_AE & 0.916 $\pm$ 0.0011 & 0.930 $\pm$ 0.0072 & 0.742 $\pm$ 0.0027 & 0.866 $\pm$ 0.0026 & 0.900 $\pm$ 0.0015 \\
		CenGCN\_WE & 0.932 $\pm$ 0.0013 & 0.973 $\pm$ 0.0053 & 0.717 $\pm$ 0.0028 & 0.892 $\pm$ 0.0017 & 0.895 $\pm$ 0.0028 \\
		CenGCN\_IE & 0.912 $\pm$ 0.0017 & 0.971 $\pm$ 0.0018 & 0.870 $\pm$ 0.0021 & 0.902 $\pm$ 0.0031 & 0.877 $\pm$ 0.0043 \\
		\bottomrule
	\end{tabular}
\end{table*}

\begin{table*}[htbp]
	
	\setlength\tabcolsep{9pt}
	\centering
	\caption{ The AUC score of link prediction.} \label{table:link2}
	\begin{tabular}{c|c|c|c|c|c}
		\toprule
		Algorithm &  Facebook & Twitter & Gplus & Youtube & LiveJournal\\
		\midrule
		GCN\_Cheby & 0.672 $\pm$ 0.0126 & 0.842 $\pm$ 0.0013 & 0.725 $\pm$ 0.0091 & 0.676 $\pm$ 0.0026 & 0.759 $\pm$ 0.0199 \\
		GCNs & 0.809 $\pm$ 0.0060 & 0.729 $\pm$ 0.0028 & 0.711 $\pm$ 0.0019 & 0.578 $\pm$ 0.0087 & 0.711 $\pm$ 0.0019 \\
		GATs & 0.633 $\pm$ 0.0041 & 0.852 $\pm$ 0.0065 & 0.558 $\pm$ 0.0057 & 0.685 $\pm$ 0.0091 & 0.757 $\pm$ 0.0179 \\
		DGI & 0.723 $\pm$ 0.0081 & 0.862 $\pm$ 0.0016 & 0.678 $\pm$ 0.0031 & 0.613 $\pm$ 0.0022 & 0.621 $\pm$ 0.0041 \\
		H-GCN & 0.708 $\pm$ 0.0180 & 0.564 $\pm$ 0.0206 & 0.601 $\pm$ 0.0204 & 0.656 $\pm$ 0.0145 & 0.739 $\pm$ 0.0024 \\
		DPSW & 0.767 $\pm$ 0. & 0.581 $\pm$ 0. & 0.797 $\pm$ 0. & 0.714 $\pm$ 0. & 0.753 $\pm$ 0. \\
		\midrule
		CenGCN\_D & \textbf{0.892 $\pm$ 0.0030} & \textbf{0.873 $\pm$ 0.0095} & \textbf{0.801 $\pm$ 0.0022} & \textbf{0.731 $\pm$ 0.0114} & 0.848 $\pm$ 0.0089 \\
		CenGCN\_TD & 0.854 $\pm$ 0.0017 & 0.857 $\pm$ 0.0053 & 0.787 $\pm$ 0.0049 & 0.718 $\pm$ 0.0063 & 0.850 $\pm$ 0.0011 \\
		CenGCN\_AD & 0.885 $\pm$ 0.0098 & 0.855 $\pm$ 0.0026 & 0.775 $\pm$ 0.0026 & 0.713 $\pm$ 0.0298 & 0.837 $\pm$ 0.0245 \\
		CenGCN\_WD & 0.884 $\pm$ 0.0335 & 0.850 $\pm$ 0.0154 & 0.796 $\pm$ 0.0057 & 0.728 $\pm$ 0.0067 & 0.831 $\pm$ 0.0013 \\
		CenGCN\_ID & 0.882 $\pm$ 0.0025 & 0.847 $\pm$ 0.0129 & 0.699 $\pm$ 0.0071 & 0.713 $\pm$ 0.0108 & 0.828 $\pm$ 0.0019 \\
		\midrule
		CenGCN\_E & 0.891 $\pm$ 0.0014 & 0.871 $\pm$ 0.0058 & 0.769 $\pm$ 0.0191 & 0.727 $\pm$ 0.0149 & \textbf{0.853 $\pm$ 0.0014} \\
		CenGCN\_TE & 0.887 $\pm$ 0.0023 & 0.858 $\pm$ 0.0017 & 0.753 $\pm$ 0.0057 & 0.715 $\pm$ 0.0138 & 0.850 $\pm$ 0.0016 \\
		CenGCN\_AE & 0.868 $\pm$ 0.0012 & 0.856 $\pm$ 0.0069 & 0.746 $\pm$ 0.0024 & 0.756 $\pm$ 0.0114 & 0.841 $\pm$ 0.0015 \\
		CenGCN\_WE & 0.808 $\pm$ 0.0154 & 0.861 $\pm$ 0.0025 & 0.742 $\pm$ 0.0204 & 0.698 $\pm$ 0.0226 & 0.848 $\pm$ 0.0011 \\
		CenGCN\_IE & 0.840 $\pm$ 0.0438 & 0.856 $\pm$ 0.0048 & 0.776 $\pm$ 0.0042 & 0.681 $\pm$ 0.0077 & 0.842 $\pm$ 0.0067 \\
		\bottomrule
	\end{tabular}
\end{table*}
%\vspace{10 mm}

\begin{IEEEbiography}[{\includegraphics[width=1in,height=1.5in,clip,keepaspectratio]{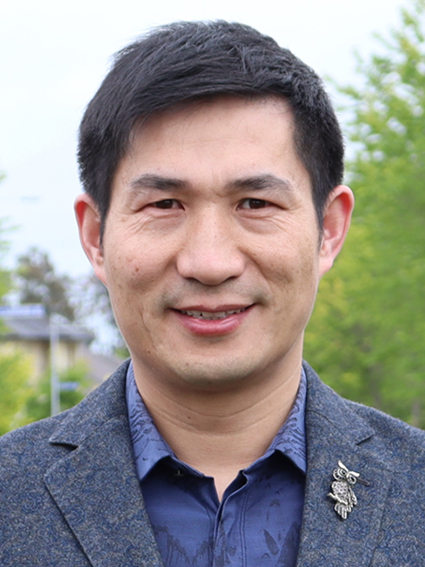}}]{Feng Xia} (M'07-SM'12) received the BSc and PhD degrees from Zhejiang University, Hangzhou, China. He was Full Professor and Associate Dean (Research) in School of Software, Dalian University of Technology, China. He is Associate Professor and former Discipline Leader (IT) in School of Engineering, IT and Physical Sciences, Federation University Australia. Dr. Xia has published 2 books and over 300 scientific papers in international journals and conferences. His research interests include data science, artificial intelligence, graph learning, and systems engineering. He is a Senior Member of IEEE and ACM. 
\end{IEEEbiography}
%\vspace{-10 mm}
\begin{IEEEbiography}[{\includegraphics[width=1in,height=1.5in,clip,keepaspectratio]{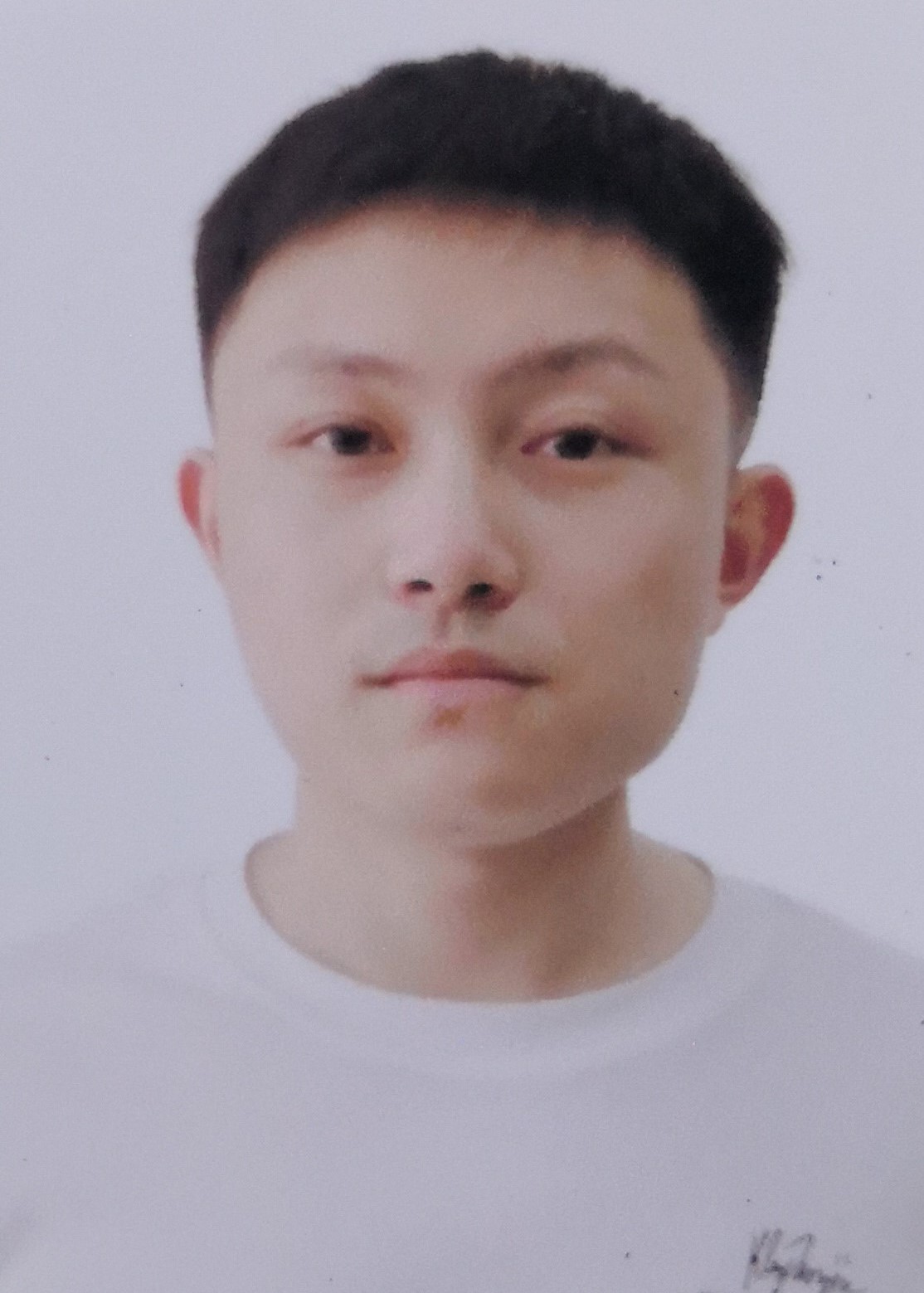}}]{Lei Wang} received the BSc degree in software engineering from Dalian University of Technology, China, in 2018. He is currently working toward the master degree in the School of Software, Dalian University of Technology, China. His research interests include data mining, analysis of complex networks, and machine learning.	
\end{IEEEbiography}
%\vspace{-10 mm}
\begin{IEEEbiography}[{\includegraphics[width=1in,height=1.5in,clip,keepaspectratio]{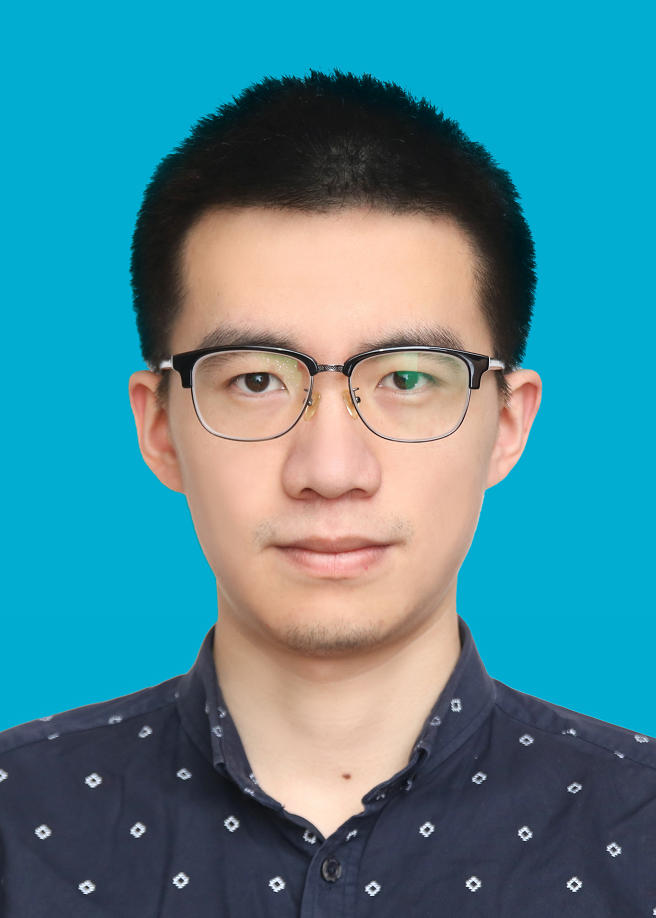}}]{Tao Tang} received the Bachelor Degree from Chengdu College, University of Electronic Science and Technology of China, Chengdu, China in 2019. He is currently pursuing the Ph.D. degree in School of Engineering, IT and Physical Sciences, Federation University Australia. His research interests include data science, recommender systems, and graph learning.
\end{IEEEbiography}
%\vspace{-10 mm}
\begin{IEEEbiography}[{\includegraphics[width=1in,height=1.5in,clip,keepaspectratio]{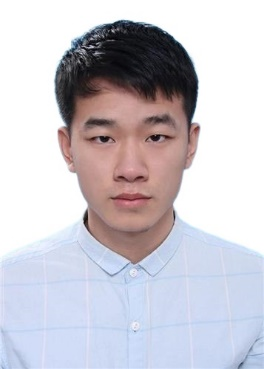}}]{Xin Chen} received the B.Sc. degree in information security from Harbin Engineering University, Harbin, China, in 2020. He is currently pursuing the master’s degree in the School of Software, Dalian University of Technology, China. His research interests include graph learning, urban science, and social computing.	
\end{IEEEbiography}
%\vspace{-10 mm}
\begin{IEEEbiography}[{\includegraphics[width=1in,height=1.5in,clip,keepaspectratio]{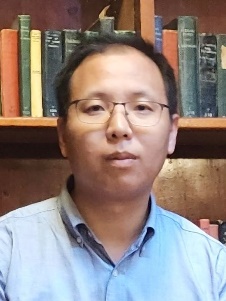}}]{Xiangjie Kong} (M'13-SM'17) received the B.Sc. and Ph.D. degrees from Zhejiang University, Hangzhou, China. He is currently a Professor with College of Computer Science and Technology, Zhejiang University of Technology. Previously, he was an Associate Professor with the School of Software, Dalian University of Technology, China. He has published over 160 scientific papers in international journals and conferences. His research interests include network science, data science, and computational social science. He is a Senior Member of IEEE.	
\end{IEEEbiography}
%\vspace{-80 mm}
\begin{IEEEbiography}[{\includegraphics[width=1in,height=1.5in,clip,keepaspectratio]{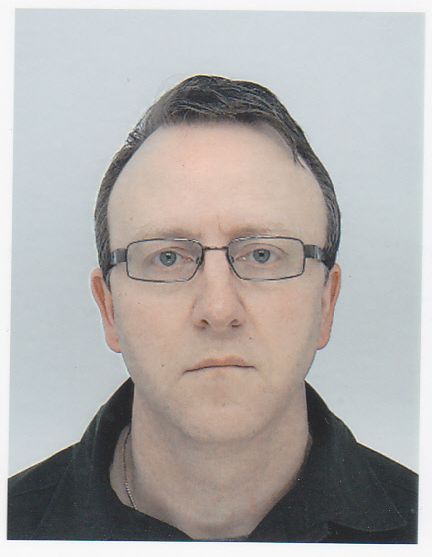}}]{Giles Oatley} received his PhD in Artificial Intelligence in 2000, and his highest position in the UK was Reader (Associate Professor) in Intelligent Systems at Cardiff Metropolitan University, before continuing his academic career in Australia since 2016. For over 20 years he has researched in data mining with particular emphasis on crime informatics, the resultant analyses often embedded in decision support systems. He is a Fellow and Chartered IT Professional with the BCS, The Chartered Institute for IT, and Charted Professional with the ACS (Australian Computing Society).	
\end{IEEEbiography}
\vspace{-160 mm}
\begin{IEEEbiography}[{\includegraphics[width=1in,height=1.5in,clip,keepaspectratio]{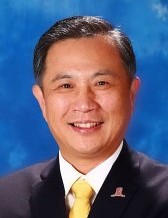}}]{Irwin King} (F'19) received the B.Sc. degree in engineering and applied science from the California Institute of Technology, Pasadena, CA, USA, and the M.Sc. and Ph.D. degrees in computer science from the University of Southern California, Los Angeles, CA. He is currently a Professor at the Department of Computer Science and Engineering, and a former Associate Dean (Education), Faculty of Engineering at The Chinese University of Hong Kong. His research interests include machine learning, social computing, web intelligence, data mining, and multimedia information processing. In these research areas, he has over 210 technical publications in journals and conferences. He is a Fellow of IEEE.	
\end{IEEEbiography}

%\end{IEEEbiography}


%\newpage



%\vfill


\end{document}